\documentclass[opre,nonblindrev]{informs3}

\OneAndAHalfSpacedXI

\RequirePackage[numbers]{natbib}
\RequirePackage[colorlinks,citecolor=blue,linkcolor=blue,urlcolor=blue,hypertexnames=false]{hyperref}
\RequirePackage{graphicx}
\usepackage{float}
\usepackage[font=footnotesize]{caption}
\usepackage{subcaption}
\usepackage{pdfpages}

\newcommand{\floor}[1]{\left\lfloor #1 \right\rfloor}
\newcommand\abs[1]{\left|#1\right|}
\def\indic#1{\mathbb{I}\left({#1}\right)}
\def\E{\mathbb{E}}
\def\P{\mathbb{P}}
\providecommand{\argmax}{\mathop\mathrm{arg max}}

\usepackage{natbib}
 \bibpunct[, ]{(}{)}{,}{a}{}{,}%
 \def\bibfont{\small}%
\TheoremsNumberedThrough     
\ECRepeatTheorems

\EquationsNumberedThrough    

\begin{document}



\RUNTITLE{The Typical Behavior of Bandit Algorithms}

\TITLE{The Typical Behavior of Bandit Algorithms}

\ARTICLEAUTHORS{%
\AUTHOR{Lin Fan}
\AFF{Department of Management Science and Engineering, Stanford University, Stanford, CA 94305, \EMAIL{linfan@stanford.edu}} 
\AUTHOR{Peter W. Glynn}
\AFF{Department of Management Science and Engineering, Stanford University, Stanford, CA 94305, \EMAIL{glynn@stanford.edu}}
} 

\ABSTRACT{%
We establish strong laws of large numbers and central limit theorems for the regret of two of the most popular bandit algorithms: Thompson sampling and UCB.
Here, our characterizations of the regret distribution complement the characterizations of the tail of the regret distribution recently developed in \cite{fan_etal2021b}.
The tail characterizations there are associated with atypical bandit behavior on trajectories where the optimal arm mean is under-estimated, leading to mis-identification of the optimal arm and large regret.
In contrast, our SLLN's and CLT's here describe the typical behavior and fluctuation of regret on trajectories where the optimal arm mean is properly estimated.
We find that Thompson sampling and UCB satisfy the same SLLN and CLT, with the asymptotics of both the SLLN and the (mean) centering sequence in the CLT matching the asymptotics of expected regret.
Both the mean and variance in the CLT grow at $\log(T)$ rates with the time horizon $T$.
Asymptotically as $T \to \infty$, the variability in the number of plays of each sub-optimal arm depends only on the rewards received for that arm, which indicates that each sub-optimal arm contributes independently to the overall CLT variance.
}%



\maketitle
%


\tableofcontents

\section{Introduction} \label{introduction}

The multi-armed bandit (MAB) problem has become an extremely fruitful area of both research and practice in recent decades.
Along with the widespread deployment of bandit algorithms in numerous diverse electronic applications, there has been a great deal of effort to better understand the performance of empirically successful algorithms from a theoretical perspective.
This literature, by now vast, is almost entirely focused on algorithm design principles which produce small regret in expectation.
Here, small means that expected regret grows as a constant multiple of $\log(T)$ with the time horizon $T$, as motivated by the Lai-Robbins lower bound \citep{lai_etal1985} which characterizes the minimum possible $\log(T)$ rate.

However, as highlighted by the recent work of \cite{fan_etal2021b}, it is important to consider other aspects of the regret distribution besides just the expected regret in bandit algorithm design.
It is shown there that focusing purely on expected regret minimization comes with several highly undesirable side effects.
First, algorithms with small or minimal rates of expected regret growth, including the most popular ones based on the Thompson sampling (TS) \citep{thompson_1933} and upper confidence bound (UCB) \citep{lai_etal1985,auer_etal2002} strategies, must have regret distributions with heavy (power law) tails.
These tails are essentially that of a truncated Cauchy distribution, implying that there is a quite large probability of suffering very large regret.
Second, expected regret minimization provides no control over the growth rate of higher moments of expected regret, and notably there is no control over the variability of regret.
Third, the truncated Cauchy tails cause an algorithm to suffer large expected regret (growing as $T^a$ for some $0 < a < 1$) when the bandit environment is just slightly mis-specified relative to the algorithm's design.

In this paper, we develop approximations to the regret distribution that complement those of \cite{fan_etal2021b}.
For fixed bandit environments, we show that as the time horizon $T \to \infty$, the regret of TS and UCB satisfy strong laws of large numbers (SLLN's) and central limit theorems (CLT's).
(For simplicity, we consider versions of TS and UCB designed for Gaussian rewards.)
In fact, these limit theorems for TS and UCB are the same, with the asymptotics of both the SLLN and the (mean) centering sequence in the CLT matching the asymptotics of expected regret.
Complementary to the characterizations of the regret distribution tail in \cite{fan_etal2021b}, the CLT's here describe the concentration and shape of the main probability mass of the regret distribution, centered around the expected regret.
The tail characterizations in \cite{fan_etal2021b} are obtained through changes of measure associated with trajectories where the optimal arm mean is under-estimated, thereby causing the optimal arm to be mis-identified and resulting in large regret.
Here, our SLLN's and CLT's are implicitly associated with trajectories where the mean of the optimal arm is properly estimated and the optimal arm is correctly identified.
In this sense, the tail of the regret distribution describes the atypical behavior of regret, whereas the SLLN's and CLT's describe the typical behavior and fluctuation of regret.

Some additional highlights of our results are as follows.
Both the means and variances in our CLT's grow at $\log(T)$ rates with $T$.
By analogy with the large deviations theory for sums of iid random variables, this suggests that large deviations of regret correspond to deviations from the expected regret that are of order $\log(T)$. 
(\cite{fan_etal2021b} characterize the tail of the regret distribution beyond $\log^{1+\epsilon}(T)$ for small $\epsilon > 0$.
Future work will analyze deviations on the $\log(T)$ scale.)
The variability in our CLT's is purely due to the variability of the sub-optimal arm rewards.
Asymptotically as $T \to \infty$, the number of plays of each sub-optimal arm depends only on the rewards received for that arm.
So the numbers of plays of different sub-optimal arms are asymptotically independent and contribute additively to the overall CLT variance.
Lastly, we find that the CLT becomes a better approximation to the regret distribution as the regret distribution tail is made lighter by increasing the amount of exploration performed by the algorithm.
(See Section 5 of \cite{fan_etal2021b} for a sharp trade-off between the amount of exploration performed by UCB-type algorithms and the resulting heaviness of the regret tail.)

In terms of related work, \cite{wager_etal2021} and \cite{fan_etal2021a} develop diffusion approximations for the regret of TS and related algorithms, and \cite{kalvit_etal2021} develop diffusion approximations for the regret of UCB.
The approximations of the regret distribution in these works are developed for bandit settings where the gaps between the arm means are roughly of size $1/\sqrt{T}$ for time horizon $T$.
So these distributional approximations are distinct from those in this paper, which are developed for bandit settings with fixed gaps between arm means as $T \to \infty$.
In \cite{cowan_etal2019}, SLLN's and laws of the iterated logarithm (LIL's) are developed for a version of UCB and also for algorithms based on forced arm sampling according to a predefined schedule.
(Our SLLN for UCB is adapted from that of \cite{cowan_etal2019}, but our SLLN's for TS and our CLT's for both UCB and TS are new.)

The rest of the paper is structured as follows.
In Section \ref{setup}, we provide a formal framework for the MAB problem and introduce notation.
In Section \ref{ts}, we develop SLLN's (Theorems \ref{thm1} and \ref{thm5}) and CLT's (Theorems \ref{thm2} and \ref{thm6}) for the regret of TS in two- and multi-armed settings.
Then, in Section \ref{ucb}, we develop a SLLN (Theorem \ref{thm3}) and CLT's (Theorems \ref{thm4} and \ref{thm7}) for the regret of UCB in two- and multi-armed settings.
In Sections \ref{ts} and \ref{ucb}, we work with versions of TS and UCB designed for environments with iid Gaussian rewards with variance $1$ (for simplicity), and we analyze their regret behavior when they operate in such environments (i.e., in well-specified settings).
Later, in Section \ref{mis}, we develop SLLN's and CLT's (Propositions \ref{prop1} and \ref{prop2}) for the regret of TS and UCB in possibly mis-specified settings.
In such settings, we work with versions of TS and UCB designed for iid Gaussian rewards with a specified variance $\sigma^2 > 0$, but we analyze their regret behavior in environments with essentially arbitrary reward distributions.
(In these mis-specified settings, we still assume that the rewards are iid for simplicity, but our technical arguments can be adapted to accommodate rewards evolving as stochastic processes.)
Finally, we examine the validity of the CLT's over finite time horizons through numerical simulations in Section \ref{numerical}.


\subsection{Model and Preliminaries} \label{setup}


A $K$-armed MAB evolves within a bandit environment $\nu = (P_1,\dots,P_K)$, where each $P_k$ is a distribution on $\mathbb{R}$.
At time $t$, the decision-maker selects an arm $A(t) \in [K] = \{1,\dots,K\}$ to play.
The conditional distribution of $A(t)$ given $A(1),Y(1),\dots,A(t-1),Y(t-1)$ is $\pi_t(\cdot \mid A(1),Y(1),\dots,A(t-1),Y(t-1))$, where $\pi = (\pi_t, t \ge 1)$ is a sequence of probability kernels, which constitutes the bandit algorithm (with $\pi_t$ defined on $([K] \times \mathbb{R})^t \times 2^{[K]}$).
Upon selecting the arm $A(t)$, a reward $Y(t)$ from arm $A(t)$ is received as feedback. 
The conditional distribution of $Y(t)$ given $A(1),Y(1),\dots,A(t-1),Y(t-1),A(t)$ is $P_{A(t)}(\cdot)$.
We write $X_k(t)$ to denote the reward received when arm $k$ is played for the $t$-th instance, so that $Y(t) = X_{A(t)}(N_{A(t)}(t))$, where $N_k(t) = \sum_{i=1}^t \indic{A(i) = k}$ denotes the number of plays of arm $k$ up to and including time $t$.
For each arm $k$, corresponding to $N_k(t)$, we use $\mathcal{T}_k(j)$ to denote the time of the $j$-th play of arm $k$, and we use
\begin{align*}
    \tau_k(j) = \mathcal{T}_k(j+1) - \mathcal{T}_k(j) 
\end{align*}
to denote the time in between the $j$-th and $(j+1)$-th plays of arm $k$.
At time $t$, the filtration for the bandit algorithms studied in this paper is given by
\begin{align*}
    \mathcal{F}_t = \{A(1),\dots,A(t), \; X_k(1),\dots,X_k(N_k(t)), \; 1 \le k \le K\}.
\end{align*}


For any time $n$, the interaction between the algorithm $\pi$ and the environment $\nu$ induces a unique probability $\P_{\nu \pi}(\cdot)$ on $([K] \times \mathbb{R})^\infty$ for which
\begin{align*}
    \P_{\nu \pi}(A(1) = a_1, Y(1) \in dy_1,\dots, A(n) = a_n, Y(n) \in dy_n) = \prod_{t=1}^n \pi_t(a_t \; | \; a_1,y_1,\dots,a_{t-1},y_{t-1}) P_{a_t}(dy_t). 
\end{align*}
Throughout the paper, all expectations and probabilities will be taken with respect to $\P_{\nu \pi}(\cdot)$.
The particular environment $\nu$ and algorithm $\pi$ under consideration will be clear from the context, and we will not write it explicitly.

The performance of an algorithm $\pi$ is measured by the (pseudo-)regret (at time $t$):
\begin{align*}
    R(t) = \sum_{k \ne k^*} N_k(t)\Delta_k, 
\end{align*}
where $\Delta_k = \mu_{k^*} - \mu_k$, $k^*$ is the optimal arm, and for any arm $k'$, $\mu_{k'}$ is the mean of its reward distribution.
(We will always assume the optimal arm is unique for technical simplicity.)
The goal in most settings is to find an algorithm $\pi$ which minimizes the expected regret $\E[R(t)]$ as $t \to \infty$.




\section{Analysis of Thompson Sampling} \label{ts}

In this section, we analyze a version of TS that is designed for iid Gaussian rewards with variance $1$.
We assume that the actual arm rewards are also iid Gaussian with variance $1$, i.e., TS is operating in a well-specified environment.
For modifications and consideration of model mis-specification, see Section \ref{mis}.

Asymptotically, the prior on the arm means used in TS does not matter, so we put a $N(0,1)$ prior on all arm means for simplicity.
Given $\mathcal{F}_{t}$ (the information collected up to and including time $t$), at time $t+1$ TS generates one sample from the posterior distribution of the mean for each arm, and then plays the arm with the highest sampled mean.
This can be implemented by generating exogenous $N(0,1)$ random variables $Z_k(t+1)$ for each arm $k$, and then playing the arm:
\begin{align*}
    A(t+1) = \argmax_{k \in [K]} \left\{ \widehat{\mu}_k(N_k(t)) + \frac{Z_k(t+1)}{\sqrt{1+N_k(t)}} \right\},
\end{align*}
where $\widehat{\mu}_k(n) = \frac{1}{1+n} \sum_{j=1}^n X_k(j)$.
As shown in \cite{korda_etal2013}, the expected regret for well-specified Gaussian TS satisfies:
\begin{align}
    \lim_{T \to \infty} \frac{\E[R(T)]}{\log(T)} = \sum_{k \ne k^*} \frac{2}{\Delta_k}. \label{ts_limit}
\end{align}
(A Jeffrey's prior is used in \cite{korda_etal2013} to derive a more general result that applies to any exponential family reward distribution.)

To develop our SLLN's and CLT's, we analyze the times $(\mathcal{T}_k(j), j \ge 1)$ during which each sub-optimal arm $k$ is played.
TS does not stop playing sub-optimal arms for any time horizon, and each sub-optimal arm is played roughly $O(\log(T))$ times by time $T$.
Thus, the spacing between the $\mathcal{T}_k(j)$ should increase exponentially with $j$.
The $\log(\mathcal{T}_k(j))$ should then be on a linear scale and satisfy SLLN's and CLT's.
Using the basic identities from renewal theory, we can obtain corresponding limit theorems for the $N_k(T)$.

To analyze the $\mathcal{T}_k(j)$, we consider probabilities of playing sub-optimal arms, as well as approximations to such probabilities.
For each sub-optimal arm $k$, define:
\begin{align}
p_k(N_k(t),t+1) & = \P\left( \widehat{\mu}_k(N_k(t)) + \frac{Z_k(t+1)}{\sqrt{1+N_k(t)}} > \widehat{\mu}_{k^*}(N_{k^*}(t)) + \frac{Z_{k^*}(t+1)}{\sqrt{1+N_{k^*}(t)}} \; \biggl| \; \mathcal{F}_{t} \right) \nonumber \\
& = \P\left( V_k(t+1) < B_k(N_k(t),t+1) \; \biggl| \; \mathcal{F}_{t} \right), \label{I1}
\end{align}
where the $V_k(t+1) = 1-\Phi(Z_k(t+1))$ are distributed according to $\text{Unif}(0,1)$, and we define
\begin{align}
    B_k(N_k(t),t+1) = 1- \Phi\left(\sqrt{1+N_k(t)} \left( \widehat{\mu}_{k^*}(N_{k^*}(t)) - \widehat{\mu}_k(N_k(t)) + \frac{Z_{k^*}(t+1)}{\sqrt{1+N_{k^*}(t)}} \right) \right). \label{I2}
\end{align}
We use a coupling setup involving the randomization variables $V_k(t+1)$ for each sub-optimal arm $k$.
For each such $k$, and each $j \ge 1$, let $(\xi_k(j,i), i \ge 1)$ be an independent exogenous sequence of $\text{Unif}(0,1)$ random variables such that 
\begin{align}
    \xi_k(j,i) = V_k(\mathcal{T}_k(j)+i), \qquad 1 \le i \le \mathcal{T}_k(j+1)-\mathcal{T}_k(j). \label{xi_k}
\end{align}
To obtain SLLN's, we use the following approximations to the $p_k$ and $\tau_k$:
\begin{align}
& \hspace{-1.7cm} \widetilde{p}_k(j) = \exp\left(-j\frac{\Delta_k^2}{2}\right) \label{I3} \\
& \hspace{-1.7cm} \widetilde{\tau}_k(j) = \inf\{ i \ge 1 : \xi_k(j,i) < \widetilde{p}_k(j) \}. \label{I4} 
\end{align}
To obtain CLT's, we use the approximations:
\begin{align}
& \hspace{1.5cm} \widehat{p}_k(j) = \frac{1}{\sqrt{\pi j (\mu_{k^*} - \widehat{\mu}_k(j))^2}} \exp\left(-\frac{1}{2}j(\mu_{k^*} - \widehat{\mu}_k(j))^2\right) \label{I5} \\
& \hspace{1.5cm} \widehat{\tau}_k(j) = \inf\{ i \ge 1 : \xi_k(j,i) < \widehat{p}_k(j) \}. \label{I6}
\end{align}

\subsection{Strong Law of Large Numbers} \label{ts_slln}

We first show that the regret of TS satisfies a SLLN in two-armed settings. 
The limit in the SLLN matches that of expected regret in (\ref{ts_limit}).
Using the coupling setup involving the exogenous $\text{Unif}(0,1)$ random variables $\xi_k(j,i)$ satisfying (\ref{xi_k}), we are able to define a simpler process involving the quantities $\widetilde{p}_k$ and $\widetilde{\tau}_k$ (approximations to $p_k$ and $\tau_k$) defined in (\ref{I3}) and (\ref{I4}).
This simpler process approximates the dynamics of TS sufficiently well to yield a SLLN.

\begin{theorem} \label{thm1}
In two-armed bandit environments with arm mean gap $\Delta > 0$, the regret of TS satisfies the SLLN:
\begin{align}
\frac{R(T)}{\log(T)} \overset{\text{a.s.}}{\to} \frac{2}{\Delta}. \label{thm1_0}
\end{align}
\end{theorem}

\proof{Proof of Theorem \ref{thm1}.}
Without loss of generality, let arm $2$ be the sub-optimal arm.
With $\widetilde{p}_2(j)$ and $\widetilde{\tau}_2(j)$ as defined in (\ref{I3}) and (\ref{I4}), note that
\begin{align}
\log\left(\sum_{j=1}^n \widetilde{\tau}_2(j) \right) & = \log\left(\sum_{j=1}^n \exp\left(j\frac{\Delta_2^2}{2} + \log(\widetilde{\tau}_2(j) \widetilde{p}_2(j))\right) \right) \nonumber \\
& \le \max_{1 \le j \le n} \left\{ j\frac{\Delta_2^2}{2} + \log(\widetilde{\tau}_2(j) \widetilde{p}_2(j)) \right\} + \log(n) \nonumber \\
& \le n\frac{\Delta_2^2}{2} +  \max_{1 \le j \le n} \log(\widetilde{\tau}_2(j) \widetilde{p}_2(j)) + \log(n), \label{S1E1}
\end{align}
where the first inequality is due Lemma \ref{lem00}.
Similarly,
\begin{align}
\log\left(\sum_{j=1}^n \widetilde{\tau}_2(j) \right) & \ge \max_{1 \le j \le n} \left\{ j \frac{\Delta_2^2}{2} + \log(\widetilde{\tau}_2(j) \widetilde{p}_2(j)) \right\} \nonumber \\
& \ge n\frac{\Delta_2^2}{2} + \log(\widetilde{\tau}_2(n) \widetilde{p}_2(n)). \label{S1E2}
\end{align}
Using Lemma \ref{lem0},
\begin{align}
\frac{\max_{1 \le j \le n} \log(\widetilde{\tau}_2(j) \widetilde{p}_2(j))}{n} \overset{\text{a.s.}}{\to} 0. \label{S1E3}
\end{align}
Then, (\ref{S1E1})-(\ref{S1E3}) together yield
\begin{align*}
\frac{\log\left(\sum_{j=1}^n \widetilde{\tau}_2(j) \right)}{n} \overset{\text{a.s.}}{\to} \frac{\Delta_2^2}{2}.
\end{align*}
So, the key to obtaining a SLLN for $N_2(t)$ is to establish that
\begin{align*}
\frac{\log\left(\sum_{j=1}^n \tau_2(j) \right)}{n} - \frac{\log\left(\sum_{j=1}^n \widetilde{\tau}_2(j) \right)}{n} \overset{\text{a.s.}}{\to} 0. 
\end{align*}
We establish this in Lemma \ref{lem1}.
Then, (\ref{thm1_0}) is established by the renewal theory relation:
\begin{align*}
\frac{\log\left(\sum_{j=1}^{N_2(t)} \tau_2(j) \right)}{N_2(t)} \le \frac{\log(t)}{N_2(t)} \le \frac{\log\left(\sum_{j=1}^{N_2(t)+1} \tau_2(j) \right)}{N_2(t)+1} \frac{N_2(t) + 1}{N_2(t)}
\end{align*}
\halmos
\endproof

\begin{lemma} \label{lem1}
Using TS in two-armed bandit environments in which arm $2$ is sub-optimal,
\begin{align}
\frac{\log\left(\sum_{j=1}^n \tau_2(j) \right)}{n} - \frac{\log\left(\sum_{j=1}^n \widetilde{\tau}_2(j) \right)}{n} \overset{\text{a.s.}}{\to} 0 \label{lem1_0}
\end{align}
\end{lemma}

\proof{Proof of Lemma \ref{lem1}.}
Let $\epsilon \in (0,\frac{\Delta_2^2}{2})$ and define
\begin{align}
& \widetilde{p}^{\;+}_2(j) = \exp\left(-j\left(\frac{\Delta_2^2}{2} + \epsilon\right)\right) \label{ptilde+} \\
& \widetilde{\tau}^{\;+}_2(j) = \inf\left\{ i \ge 1 : \xi_2(j,i) < \widetilde{p}^{\;+}_2(j) \right\} \nonumber \\
& \widetilde{p}^{\;-}_2(j) = 2\exp\left(-j\left(\frac{\Delta_2^2}{2} - \epsilon\right)\right) \label{ptilde-} \\
& \widetilde{\tau}^{\;-}_2(j) = \inf\left\{ i \ge 1 : \xi_2(j,i) < \widetilde{p}^{\;-}_2(j) \right\}. \nonumber
\end{align}
Since $\widetilde{\tau}_2(j)$, $\widetilde{\tau}^{\;+}_2(j)$ and $\widetilde{\tau}^{\;-}_2(j)$ are all defined using a common set of random variables $\xi_2(j,i)$, we have almost surely for all $j$,
\begin{align}
\widetilde{\tau}^{\;-}_2(j) \le \widetilde{\tau}_2(j) \le \widetilde{\tau}^{\;+}_2(j). \label{S1E5}
\end{align}
We claim that, almost surely, for sufficiently large $j$,
\begin{align}
\widetilde{\tau}^{\;-}_2(j) \le \tau_2(j) \le \widetilde{\tau}^{\;+}_2(j). \label{S1E6}
\end{align}
This follows from
\begin{align*}
\widetilde{p}^{\;+}_2(N_2(t)) < p_2(N_2(t),t+1) < \widetilde{p}^{\;-}_2(N_2(t)), 
\end{align*}
which is established in Lemma \ref{lem2}.
Because of (\ref{S1E5}) and (\ref{S1E6}), we have, almost surely,
\begin{align}
\limsup_{n \to \infty} \abs{\frac{\log\left(\sum_{j=1}^n \tau_2(j) \right)}{n} - \frac{\log\left(\sum_{j=1}^n \widetilde{\tau}_2(j) \right)}{n}} \le \limsup_{n \to \infty} \left( \frac{\log\left(\sum_{j=1}^n \widetilde{\tau}^{\;+}_2(j) \right)}{n} - \frac{\log\left(\sum_{j=1}^n \widetilde{\tau}^{\;-}_2(j) \right)}{n} \right). \label{S1E7}
\end{align}
We now show that the right-hand side of (\ref{S1E7}) is almost surely negligible.
Note that
\begin{align}
0 & \le \log\left(\sum_{j=1}^n \widetilde{\tau}^{\;+}_2(j) \right) - \log\left(\sum_{j=1}^n \widetilde{\tau}^{\;-}_2(j) \right) \nonumber \\
& \le \max_{1 \le j \le n} \left\{ j\left(\frac{\Delta_2^2}{2} + \epsilon\right) + \log\left(\widetilde{\tau}^{\;+}_2(j) \widetilde{p}^{\;+}_2(j)\right) \right\} + \log(n) - \max_{1 \le j \le n} \left\{ j\left(\frac{\Delta_2^2}{2} - \epsilon\right) + \log\left(2^{-1} \widetilde{\tau}^{\;-}_2(j) \widetilde{p}^{\;-}_2(j)\right) \right\} \nonumber \\
& \le n\left(\frac{\Delta_2^2}{2} + \epsilon\right) + \max_{1 \le j \le n} \log\left(\widetilde{\tau}^{\;+}_2(j) \widetilde{p}^{\;+}_2(j)\right) + \log(n) - n\left(\frac{\Delta_2^2}{2} - \epsilon\right) - \log\left(\widetilde{\tau}^{\;-}_2(n) \widetilde{p}^{\;-}_2(n)\right) + \log(2), \label{S1E8}
\end{align}
where the first inequality holds by the definition of $\widetilde{\tau}^{\;+}_2(j)$ and $\widetilde{\tau}^{\;-}_2(j)$ (see (\ref{S1E5})), and the second inequality is due Lemma \ref{lem00}.
Using Lemma \ref{lem0}, we have
\begin{align}
& \frac{\max_{1 \le j \le n} \log\left(\widetilde{\tau}^{\;+}_2(j) \widetilde{p}^{\;+}_2(j)\right)}{n} \overset{\text{a.s.}}{\to} 0 \label{S1E9} \\
& \frac{\max_{1 \le j \le n} \log\left(\widetilde{\tau}^{\;-}_2(j) \widetilde{p}^{\;-}_2(j)\right)}{n} \overset{\text{a.s.}}{\to} 0. \label{S1E10}
\end{align}
Putting together (\ref{S1E8})-(\ref{S1E10}),
\begin{align*}
\limsup_{n \to \infty} \left(\frac{\log\left(\sum_{j=1}^n \widetilde{\tau}^{\;+}_2(j) \right)}{n} - \frac{\log\left(\sum_{j=1}^n \widetilde{\tau}^{\;-}_2(j) \right)}{n}\right) \le 2\epsilon,
\end{align*}
and so, together with (\ref{S1E7}), we have, almost surely,
\begin{align*}
\limsup_{n \to \infty} \abs{\frac{\log\left(\sum_{j=1}^n \tau_2(j) \right)}{n} - \frac{\log\left(\sum_{j=1}^n \widetilde{\tau}_2(j) \right)}{n}} \le 2\epsilon.
\end{align*}
Sending $\epsilon \downarrow 0$ yields (\ref{lem1_0}).
\halmos
\endproof

\begin{lemma} \label{lem2}
Using TS in two-armed bandit environments in which arm $2$ is sub-optimal, with $\widetilde{p}_2^{\;+}$ and $\widetilde{p}_2^{\;_-}$ as defined in (\ref{ptilde+}) and (\ref{ptilde-}), for sufficiently large $t$,
\begin{align}
\widetilde{p}_2^{\;+}(N_2(t)) < p_2(N_2(t),t+1) < \widetilde{p}_2^{\;-}(N_2(t)). \label{lem2_0}
\end{align}
\end{lemma}

\proof{Proof of Lemma \ref{lem2}.}
In (\ref{I1}) and (\ref{I2}), we provide control over the term 
\begin{align*}
    B_2(N_2(t),t+1) = 1- \Phi\left(\sqrt{1+N_2(t)} \left( \widehat{\mu}_1(N_1(t)) - \widehat{\mu}_2(N_2(t)) + \frac{Z_1(t+1)}{\sqrt{1+N_1(t)}} \right) \right).
\end{align*}
By Theorem 1 of \cite{may_etal2012}, $N_2(t)/N_1(t) \overset{\text{a.s.}}{\to} 0$ and $N_2(t) \overset{\text{a.s.}}{\to} \infty$.
Let $\epsilon' > 0$.
Almost surely, for $t$ sufficiently large, 
\begin{align*}
\abs{\widehat{\mu}_1(N_1(t)) - \widehat{\mu}_2(N_2(t)) - \Delta_2} \le \frac{\epsilon'}{2}.    
\end{align*}
Consider the event
\begin{align*}
    D_{t+1} = \left\{ \abs{\frac{Z_1(t+1)}{\sqrt{1+N_1(t)}}} \le \frac{\epsilon'}{2} \right\}.
\end{align*}
Then almost surely, for $t$ sufficiently large,
\begin{align*}
    1-\Phi\left(\sqrt{N_2(t)}(\Delta_2 + \epsilon')\right) & \le \P\left( V_2(t+1) < B_2(N_2(t),t+1), \; D_{t+1} \mid \mathcal{F}_t \right) \\
    & \le \P\left( V_2(t+1) < B_2(N_2(t),t+1) \mid \mathcal{F}_t \right) \quad \Bigl( = p_2(N_2(t),t+1) \Bigr) \\
    & \le \P\left( V_2(t+1) < B_2(N_2(t),t+1), \; D_{t+1} \mid \mathcal{F}_t \right) + \P\left(D_{t+1}^c \right) \\
    & \le 2\left(1-\Phi\left(\sqrt{N_2(t)}(\Delta_2 - \epsilon')\right)\right),
\end{align*}
where the last inequality is due to $N_2(t)/N_1(t) \overset{\text{a.s.}}{\to} 0$, and so $\P(D_{t+1}^c)$ is asymptotically negligible compared to $\P\left( V_2(t+1) < B_2(N_2(t),t+1), \; D_{t+1} \mid \mathcal{F}_t \right)$.
Then, (\ref{lem2_0}) is established by taking $\epsilon' > 0$ sufficiently small and applying Lemma \ref{lem000}.
\halmos
\endproof

\subsection{Central Limit Theorem} \label{ts_clt}

We now show that the regret of TS satisfies a CLT in two-armed settings.
The (mean) centering in the CLT matches the asymptotics of the SLLN in Theorem \ref{thm1} (and also that of expected regret in (\ref{ts_limit})).
To prove the CLT, we use an approach similar to that used to prove the SLLN, but we use a process involving the refined quantities $\widehat{p}_k$ and $\widehat{\tau}_k$ (approximations to $p_k$ and $\tau_k$) defined in (\ref{I5}) and (\ref{I6}).
It turns out that the variability in the CLT is purely due to the variability of the rewards of the sub-optimal arm.
This is reasonable in light of the fact that the optimal arm is played much more than the sub-optimal arm, and so its sample mean is much more concentrated around its true mean.

\begin{theorem} \label{thm2}
In two-armed bandit environments with arm mean gap $\Delta > 0$, the regret of TS satisfies the CLT:
\begin{align}
\frac{R(T) - \frac{2}{\Delta} \log(T)}{\frac{2}{\Delta}\sqrt{2\log(T)}} \Rightarrow N(0,1). \label{thm2_0}
\end{align}
\end{theorem}

\proof{Proof of Theorem \ref{thm2}.}
Without loss of generality, let arm $2$ be the sub-optimal arm.
With $\widehat{p}_2(j)$ and $\widehat{\tau}_2(j)$ as defined in (\ref{I5}) and (\ref{I6}), note that
\begin{align}
\log\left(\sum_{j=1}^n \widehat{\tau}_2(j) \right) & = \log\left(\sum_{j=1}^n \exp\left(\frac{1}{2}j(\mu_1 - \widehat{\mu}_2(j))^2 + \log\left(\sqrt{\pi j (\mu_1 - \widehat{\mu}_2(j))^2}\right) + \log(\widehat{\tau}_2(j) \widehat{p}_2(j))\right) \right) \nonumber \\
& \le \max_{1 \le j \le n} \left\{ \frac{1}{2}j(\mu_1 - \widehat{\mu}_2(j))^2 + \log\left(\sqrt{\pi j (\mu_1 - \widehat{\mu}_2(j))^2}\right) + \log(\widehat{\tau}_2(j) \widehat{p}_2(j)) \right\} + \log(n) \nonumber \\
& \le \max_{1 \le j \le n} \frac{1}{2}j(\mu_1 - \widehat{\mu}_2(j))^2 + \max_{1 \le j \le n} \log\left(\sqrt{\pi j (\mu_1 - \widehat{\mu}_2(j))^2}\right) + \max_{1 \le j \le n} \log(\widehat{\tau}_2(j) \widehat{p}_2(j)) + \log(n), \label{ts1}
\end{align}
where the first inequality is due to Lemma \ref{lem00}.
Similarly,
\begin{align}
\log\left(\sum_{j=1}^n \widehat{\tau}_2(j) \right) & \ge \max_{1 \le j \le n} \left\{ \frac{1}{2}j(\mu_1 - \widehat{\mu}_2(j))^2 + \log\left(\sqrt{\pi j (\mu_1 - \widehat{\mu}_2(j))^2}\right) + \log(\widehat{\tau}_2(j) \widehat{p}_2(j)) \right\} \nonumber \\
& \ge \frac{1}{2} n(\mu_1 - \widehat{\mu}_2(n))^2 + \log\left(\sqrt{\pi n (\mu_1 - \widehat{\mu}_2(n))^2}\right) + \log(\widehat{\tau}_2(n) \widehat{p}_2(n)). \label{ts2}
\end{align}
Since $\widehat{\mu}_2(j) \overset{\text{a.s.}}{\to} \mu_2$ by the SLLN, it is straightforward to see that
\begin{align}
\frac{\max_{1 \le j \le n} \log\left(\sqrt{\pi j (\mu_1 - \widehat{\mu}_2(j))^2}\right)}{\sqrt{n}} \overset{\text{a.s.}}{\to} 0. \label{ts3}
\end{align}
Also, using Lemma \ref{lem0},
\begin{align}
\frac{\max_{1 \le j \le n} \log(\widehat{\tau}_2(j) \widehat{p}_2(j))}{\sqrt{n}} \overset{\text{a.s.}}{\to} 0. \label{ts4}
\end{align}
For any $j$,
\begin{align}
\frac{1}{2} j \left(\mu_1 - \widehat{\mu}_2(j)\right)^2 & = \Delta_2 \sum_{i=1}^j \left(Y_2(i) + \frac{\Delta_2}{2}\right) + \left(\frac{1}{\sqrt{2j}} \sum_{i=1}^j Y_2(i) \right)^2, \label{ts5}
\end{align}
where $Y_2(i) = -(X_2(i) - \mu_2)$ is an independent sequence of $N(0,1)$ random variables.
By the LIL, 
\begin{align}
\frac{1}{\sqrt{n}} \max_{1 \le j \le n} (\log \log j) \left(\frac{1}{\sqrt{2 j \log \log j}} \sum_{i=1}^j Y_2(i) \right)^2 \overset{\text{a.s.}}{\to} 0. \label{ts6}
\end{align}
Using (\ref{ts1}) together with (\ref{ts3})-(\ref{ts6}), we have for any $x \in \mathbb{R}$,
\begin{align}
    \liminf_{n \to \infty} \P\left( \frac{\left( \max_{1 \le j \le n} \Delta_2 \sum_{i=1}^j \left(Y_2(i) + \frac{\Delta_2}{2}\right) \right) - n\frac{\Delta_2^2}{2}}{\Delta_2\sqrt{n}} \le x \right) \le \liminf_{n \to \infty} \P\left( \frac{\log\left( \sum_{j=1}^n \widehat{\tau}_2(j) \right) - n\frac{\Delta_2^2}{2}}{\Delta_2\sqrt{n}} \le x \right). \label{ts7}
\end{align}
Using (\ref{ts2}) together with (\ref{ts3})-(\ref{ts6}), we have for any $x \in \mathbb{R}$,
\begin{align}
    \limsup_{n \to \infty} \P\left( \frac{\log\left( \sum_{j=1}^n \widehat{\tau}_2(j) \right) - n\frac{\Delta_2^2}{2}}{\Delta_2\sqrt{n}} \le x \right) \le \limsup_{n \to \infty} \P\left( \frac{\Delta_2 \sum_{i=1}^n \left(Y_2(i) + \frac{\Delta_2}{2}\right) - n\frac{\Delta_2^2}{2}}{\Delta_2\sqrt{n}} \le x \right). \label{ts8}
\end{align}
By Theorem 2.12.3 of \cite{gut_2009}, the left side of (\ref{ts7}) is $\Phi(x)$, and by the classical CLT, the right side of (\ref{ts8}) is also $\Phi(x)$.
So we have shown that
\begin{align}
    \frac{\log\left( \sum_{j=1}^n \widehat{\tau}_2(j) \right) - n\frac{\Delta_2^2}{2}}{\Delta_2\sqrt{n}} \Rightarrow N(0,1). \label{ts13}
\end{align}
So, the key to obtaining a CLT for $N_2(t)$ is to establish that 
\begin{align}
\frac{\log\left(\sum_{j=1}^n \tau_2(j) \right)}{\sqrt{n}} - \frac{\log\left(\sum_{j=1}^n \widehat{\tau}_2(j) \right)}{\sqrt{n}} \overset{\P}{\to} 0. \label{S2C1}
\end{align}
We establish this in Lemma \ref{lem3}.
Then we can apply the standard renewal process CLT argument as follows.
Let $x \in \mathbb{R}$, and define $h(t) = \floor{x \frac{2}{\Delta_2^{2}} \sqrt{2t} + \frac{2}{\Delta_2^2} t}$.
Then,
\begin{align*}
\P\left( \frac{\log\left(\sum_{j=1}^{h(t)} \tau_2(j) \right) - h(t)\frac{\Delta_2^2}{2}}{\Delta_2\sqrt{h(t)}} > -x \right) & \sim \P\left( \frac{\log\left(\sum_{j=1}^{h(t)} \tau_2(j) \right) - h(t)\frac{\Delta_2^2}{2}}{\Delta_2\sqrt{h(t)}} > \frac{t - h(t)\frac{\Delta_2^2}{2}}{\Delta_2\sqrt{h(t)}} \right) \\ 
& = \P\left( \sum_{j=1}^{h(t)} \tau_2(j) > e^t \right) \\
& = \P\left( N_2(e^t) \le h(t) \right) \\
& = \P\left(\frac{N_2(e^t) - t \frac{2}{\Delta_2^2}}{\frac{2}{\Delta_2^2} \sqrt{2t}} \le x \right).
\end{align*}
Using this together with (\ref{ts13}) and (\ref{S2C1}), (\ref{thm2_0}) is established.
\halmos
\endproof

\begin{lemma} \label{lem3}
Using TS in two-armed bandit environments in which arm $2$ is sub-optimal,
\begin{align}
\frac{\log\left(\sum_{j=1}^n \tau_2(j) \right)}{\sqrt{n}} - \frac{\log\left(\sum_{j=1}^n \widehat{\tau}_2(j) \right)}{\sqrt{n}} \overset{\P}{\to} 0. \label{lem3_0}
\end{align}
\end{lemma}

\proof{Proof of Lemma \ref{lem3}.}
Define
\begin{align}
& \widehat{p}^{\;+}_2(j) = \frac{1}{5} \widehat{p}_2(j) \label{phat+} \\
& \widehat{\tau}^{\;+}_2(j) = \inf\left\{ i \ge 1 : \xi_2(j,i) < \widetilde{p}^{\;+}_2(j) \right\}, \nonumber
\end{align}
with $\widehat{p}_2(j)$ as defined in (\ref{I5}).
Since $\widehat{\tau}_2(j)$ and $\widehat{\tau}^{\;+}_2(j)$ are all defined using a common set of random variables $\xi_2(j,i)$, we have almost surely for all $j$,
\begin{align*}
\widehat{\tau}_2(j) \le \widehat{\tau}^{\;+}_2(j). 
\end{align*}
We claim that, almost surely, for sufficiently large $j$,
\begin{align}
\widehat{\tau}_2(j) \le \tau_2(j) \le \widehat{\tau}^{\;+}_2(j). \label{lem3_2}
\end{align}
This follows from
\begin{align*}
\widehat{p}^{\;+}_2(N_2(t)) < p_2(N_2(t),t+1) < \widehat{p}_2(N_2(t)), 
\end{align*}
which is established in Lemma \ref{lem4}.

Because of (\ref{lem3_2}),
\begin{align}
     \abs{\frac{\log\left(\sum_{j=1}^n \tau_2(j) \right)}{\sqrt{n}} - \frac{\log\left(\sum_{j=1}^n \widehat{\tau}_2(j) \right)}{\sqrt{n}}} \le \abs{\frac{\log\left(\sum_{j=1}^n \widehat{\tau}^{\;+}_2(j) \right)}{\sqrt{n}} - \frac{\log\left(\sum_{j=1}^n \widehat{\tau}_2(j) \right)}{\sqrt{n}}} + o_{\text{a.s.}}(1). \label{ts10}
\end{align}
We now show that (for the right-hand side of (\ref{ts10})),
\begin{align}
\frac{\log\left(\sum_{j=1}^n \widehat{\tau}^{\;+}_2(j) \right)}{\sqrt{n}} - \frac{\log\left(\sum_{j=1}^n \widehat{\tau}_2(j) \right)}{\sqrt{n}} \overset{\P}{\to} 0. \label{ts18}
\end{align}
Similar to previous arguments,
\begin{align}
0 & \le \log\left(\sum_{j=1}^n \widehat{\tau}^{\;+}_2(j) \right) - \log\left(\sum_{j=1}^n \widehat{\tau}_2(j) \right) \nonumber \\
& \le \max_{1 \le j \le n} \left\{ \frac{1}{2}j(\mu_1 - \widehat{\mu}_2(j))^2 + \log\left(5\sqrt{\pi j (\mu_1 - \widehat{\mu}_2(j))^2}\right) + \log\left(\widehat{\tau}^{\;+}_2(j) \widehat{p}^{\;+}_2(j)\right) \right\} + \log(n) \nonumber \\
& \quad - \max_{1 \le j \le n} \left\{ \frac{1}{2}j(\mu_1 - \widehat{\mu}_2(j))^2 + \log\left(\sqrt{\pi j (\mu_1 - \widehat{\mu}_2(j))^2}\right) + \log\left(\widehat{\tau}_2(j) \widehat{p}_2(j)\right) \right\} \nonumber \\
& \le \max_{1 \le j \le n} \frac{1}{2}j(\mu_1 - \widehat{\mu}_2(j))^2 + \max_{1 \le j \le n} \log\left(5\sqrt{\pi j (\mu_1 - \widehat{\mu}_2(j))^2}\right) + \max_{1 \le j \le n} \log\left(\widehat{\tau}^{\;+}_2(j) \widehat{p}^{\;+}_2(j)\right) + \log(n) \nonumber \\ 
& \quad - \frac{1}{2}n(\mu_1 - \widehat{\mu}_2(n))^2 - \log\left(\sqrt{\pi n (\mu_1 - \widehat{\mu}_2(n))^2}\right) - \log\left(\widehat{\tau}_2(n) \widehat{p}_2(n)\right), \label{ts11}
\end{align}
where the first inequality holds by the definition of $\widehat{\tau}^{\;+}_2(j)$ and $\widehat{\tau}_2(j)$, and the second inequality is due to Lemma \ref{lem00}.
Using Lemma \ref{lem0},
\begin{align}
& \frac{\max_{1 \le j \le n} \log\left(\widehat{\tau}^{\;+}_2(j) \widehat{p}^{\;+}_2(j)\right)}{\sqrt{n}} \overset{\text{a.s.}}{\to} 0. \label{ts12}
\end{align}
Using (\ref{ts11}) and (\ref{ts12}), together with (\ref{ts3})-(\ref{ts6}), we have
\begin{align}
    0 & \le \frac{\log\left(\sum_{j=1}^n \widehat{\tau}^{\;+}_2(j) \right)}{\sqrt{n}} - \frac{\log\left(\sum_{j=1}^n \widehat{\tau}_2(j) \right)}{\sqrt{n}} \nonumber \\
    & \le \Delta_2 \frac{1}{\sqrt{n}} \left(\max_{1 \le j \le n} \sum_{i=1}^j \left(Y_2(i) + \frac{\Delta_2}{2}\right) - \sum_{i=1}^n \left(Y_2(i) + \frac{\Delta_2}{2}\right) \right) + o_{\text{a.s.}}(1). \label{ts19}
\end{align}
The random walk $\sum_{i=1}^j \left(Y_2(i) + \frac{\Delta_2}{2}\right)$ has positive drift, and so by Lemma 1.4.1 of \cite{prabhu_1998},
\begin{align}
    \frac{1}{\sqrt{n}} \left(\max_{1 \le j \le n} \sum_{i=1}^j \left(Y_2(i) + \frac{\Delta_2}{2}\right) - \sum_{i=1}^n \left(Y_2(i) + \frac{\Delta_2}{2}\right) \right) \overset{\P}{\to} 0. \label{ts20}
\end{align}
Putting together (\ref{ts19}) and (\ref{ts20}), we have established (\ref{ts18}).
Then, (\ref{lem3_0}) is established using (\ref{ts10}) and (\ref{ts18}).
\halmos
\endproof

\begin{lemma} \label{lem4}
Using TS in two-armed bandit environments in which arm $2$ is sub-optimal, with $\widehat{p}_2^{\;+}$ as defined in (\ref{phat+}), for sufficiently large $t$,
\begin{align}
\widehat{p}_2^{\;+}(N_2(t)) < p_2(N_2(t),t+1) < \widehat{p}_2(N_2(t)). \label{lem4_0}
\end{align}
\end{lemma}

\proof{Proof of Lemma \ref{lem4}.}
In (\ref{I1}) and (\ref{I2}), we provide control over the term 
\begin{align*}
    & B_2(N_2(t),t+1) \\
    & = 1 - \Phi\left(\sqrt{1+N_2(t)} \left( \widehat{\mu}_1(N_1(t)) - \widehat{\mu}_2(N_2(t)) + \frac{Z_1(t+1)}{\sqrt{1+N_1(t)}} \right) \right) \\
    & = 1 - \Phi\left( \sqrt{1+N_2(t)}\Bigl( \mu_1 - \widehat{\mu}_2(N_2(t)) \Bigr) + \sqrt{1+N_2(t)}\Bigl( \widehat{\mu}_1(N_1(t)) - \mu_1 \Bigr) + \sqrt{\frac{1+N_2(t)}{1+N_1(t)}} t^{1/4} \frac{Z_1(t+1)}{t^{1/4}} \right).
\end{align*}
From Theorem \ref{thm1},
\begin{align}
& \frac{N_2(t)}{\log(t)} \overset{\text{a.s.}}{\to} \frac{2}{\Delta_2^2} \label{lem4_1} \\
& \frac{N_1(t)}{t} \overset{\text{a.s.}}{\to} 1. \label{lem4_2}
\end{align}
Consider the event
\begin{align*}
    D_{t+1} = \left\{ \abs{\frac{Z_1(t+1)}{t^{1/4}}} \le 1 \right\}.
\end{align*}
Then almost surely, for $t$ sufficiently large,
\begin{align*}
    & 1-\Phi\left(\sqrt{1+N_2(t)}\Bigl(\mu_1 - \widehat{\mu}_2(N_2(t))\Bigr) + O_{\text{a.s.}}\left( \sqrt{\log(t)}t^{-1/4} \right) \right) \\
    & \le \P\left( V_2(t+1) < B_2(N_2(t),t+1), \; D_{t+1} \mid \mathcal{F}_t \right) \\
    & \le \P\left( V_2(t+1) < B_2(N_2(t),t+1) \mid \mathcal{F}_t \right) \quad \Bigl( = p_2(N_2(t),t+1) \Bigr) \\
    & \le \P\left( V_2(t+1) < B_2(N_2(t),t+1), \; D_{t+1} \mid \mathcal{F}_t \right) + \P\left(D_{t+1}^c \right) \\
    & \le 2\left(1-\Phi\left(\sqrt{1+N_2(t)}\Bigl(\mu_1 - \widehat{\mu}_2(N_2(t))\Bigr) + O_{\text{a.s.}}\left( \sqrt{\log(t)}t^{-1/4} \right) \right)\right),
\end{align*}
where the last inequality is due to (\ref{lem4_1}) and (\ref{lem4_2}), and so $\P(D_{t+1}^c)$ is asymptotically negligible compared to $\P\left( V_2(t+1) < B_2(N_2(t),t+1), \; D_{t+1} \mid \mathcal{F}_t \right)$.
Then, (\ref{lem4_0}) is established using Lemma \ref{lem000}.
\halmos
\endproof

\subsection{Extension to Multiple Arms} \label{ts_multi}

In this section, we extend the SLLN and CLT for the regret of TS in two-armed settings (Theorems \ref{thm1} and \ref{thm2}) to multi-armed settings.
The key to the extensions is the fact that compared to the probability of a single sub-optimal arm sampled mean exceeding that of the optimal arm, there is a much lower probability that two or more sub-optimal arm sampled means exceed that of the optimal arm.
So, effectively, each sub-optimal arm only competes with the optimal arm to be played, and the analysis in multi-armed settings reduces to that in the two-armed setting.
Moreover, $N_k(T)$ for each sub-optimal arm $k$ depends only on the rewards received for that arm.
So the $N_k(T)$ of different sub-optimal arms $k$ are independent and contribute additively to the overall CLT variance.

\begin{theorem} \label{thm5}
Using TS, for each sub-optimal arm $k$,
\begin{align}
\frac{N_k(T)}{\log(T)} \overset{\text{a.s.}}{\to} \frac{2}{\Delta_k^2}. \label{thm5_1}
\end{align}
Therefore, the regret satisfies the SLLN:
\begin{align}
\frac{R(T)}{\log(T)} \overset{\text{a.s.}}{\to} \sum_{k \ne k^*} \frac{2}{\Delta_k}. \label{thm5_2}
\end{align}
\end{theorem}

\begin{theorem} \label{thm6}
Using TS, for each sub-optimal arm $k$,
\begin{align}
\frac{N_k(T) - \frac{2}{\Delta_k^2} \log(T)}{\frac{2}{\Delta_k^2}\sqrt{2\log(T)}} \Rightarrow N(0,1). \label{thm6_1}
\end{align}
Furthermore, for different sub-optimal arms $k$, the $N_k(T)$ are asymptotically independent.
Therefore, the regret satisfies the CLT:
\begin{align}
\frac{R(T) - \sum_{k \ne k^*} \frac{2}{\Delta_k} \log(T)}{\sqrt{\sum_{k \ne k^*} \frac{8}{\Delta_k^2} \log(T)}} \Rightarrow N(0,1). \label{thm6_2}
\end{align}
\end{theorem}

\proof{Proof of Theorems \ref{thm5} and \ref{thm6}.}
Let $\delta \in (0, \min_{k'} \Delta_{k'})$.
Denote the event
\begin{align*}
    D_{t+1} = \left\{ \forall k' \ne k, k^* \; : \; \frac{Z_{k'}(t+1)}{M} < \frac{\delta}{4} \right\} \cap \left\{ \abs{\frac{Z_{k^*}(t+1)}{M}} < \frac{\delta}{4} \right\}.
\end{align*}
Set $M > 0$ sufficiently large so that $\P(D_{t+1}) \ge 1/2$ (for all $t$).
When the bandit environment involves more than two arms, the analysis can still be reduced to the two-armed case.
In particular, effectively each sub-optimal arm $k$ only competes with the optimal arm $k^*$ to be played, and the behavior of $N_k(t)$ only depends on the rewards received for that sub-optimal arm $k$.
The probability upper and lower bounds in (\ref{thm56_1})-(\ref{thm56_4}) show that each sub-optimal arm $k$ effectively only competes with the optimal arm $k^*$ to be played.
Indeed, the probabilities in (\ref{thm56_1}) and (\ref{thm56_4}) can be approximated using the same arguments from Lemmas \ref{lem2} and \ref{lem4}, leading respectively to (\ref{thm5_1}) and (\ref{thm6_1}) for each sub-optimal arm $k$.
\begin{align}
    & \P\left( \widehat{\mu}_k(N_k(t)) + \frac{Z_k(t+1)}{\sqrt{1+N_k(t)}} > \widehat{\mu}_{k^*}(N_{k^*}(t)) + \frac{\delta}{4} \; \biggl| \; \mathcal{F}_t \right) \cdot \frac{1}{2} \label{thm56_1} \\
    & \le \P\left( \widehat{\mu}_k(N_k(t)) + \frac{Z_k(t+1)}{\sqrt{1+N_k(t)}} > \widehat{\mu}_{k^*}(N_{k^*}(t)) + \frac{Z_{k^*}(t+1)}{\sqrt{1+N_{k^*}(t)}}, \; D_{t+1} \; \biggl| \; \mathcal{F}_t \right) \label{thm56_2} \\
    & \le \P\left( \widehat{\mu}_k(N_k(t)) + \frac{Z_k(t+1)}{\sqrt{1+N_k(t)}} > \max_{k' \ne k} \left\{ \widehat{\mu}_{k'}(N_{k'}(t)) + \frac{Z_{k'}(t+1)}{\sqrt{1+N_{k'}(t)}} \right\} \biggl| \; \mathcal{F}_t \right) \label{thm56_3} \\
    & \le \P\left( \widehat{\mu}_k(N_k(t)) + \frac{Z_k(t+1)}{\sqrt{1+N_k(t)}} > \widehat{\mu}_{k^*}(N_{k^*}(t)) + \frac{Z_{k^*}(t+1)}{\sqrt{1+N_{k^*}(t)}} \; \biggl| \; \mathcal{F}_t \right) \label{thm56_4}
\end{align}
For each sub-optimal arm $k$, $N_k(t)/N_{k^*}(t) \overset{\text{a.s.}}{\to} 0$ and $N_k(t) \overset{\text{a.s.}}{\to} \infty$ using Theorem 1 of \cite{may_etal2012}.
So for $t$ sufficiently large, $\sqrt{1+N_{k'}(t)} > M$ for all arms $k'$.
Using the event $D_{t+1}$ (satisfying $\P(D_{t+1}) \ge 1/2$ by construction), we obtain (\ref{thm56_2}).
To obtain (\ref{thm56_3}), note that for any sub-optimal arm $k' \ne k$, almost surely for sufficiently large $t$, on the event $D_{t+1}$,
\begin{align*}
    \widehat{\mu}_{k'}(N_{k'}(t)) + \frac{Z_{k'}(t+1)}{\sqrt{1+N_{k'}(t)}} < \mu_{k'} + \frac{\delta}{2} < \mu_{k^*} - \frac{\delta}{2} < \widehat{\mu}_{k^*}(N_{k^*}(t)) + \frac{Z_{k^*}(t+1)}{\sqrt{1+N_{k^*}(t)}}.
\end{align*}

To obtain (\ref{thm5_2}) from (\ref{thm5_1}), we only need to add up the contributions of each $N_k(t)$ to the regret $R(t)$.
To obtain (\ref{thm6_2}) from (\ref{thm6_1}), we next show that for different sub-optimal arms $k$, the $N_k(t)$ are asymptotically independent.
From the proof of Lemma \ref{lem3}, in particular using the arguments leading to (\ref{ts1}), (\ref{ts2}) and (\ref{S2C1}), we have almost surely,
\begin{align*}
    \frac{\max_{1 \le j \le n} \frac{1}{2}j(\mu_{k^*} - \widehat{\mu}_k(j))^2}{\sqrt{n}} + o_{\text{a.s.}}(1) \le \frac{\log\left(\sum_{j=1}^n \widehat{\tau}_k(j) \right)}{\sqrt{n}} \le \frac{\frac{1}{2}n(\mu_{k^*} - \widehat{\mu}_k(n))^2}{\sqrt{n}} + o_{\text{a.s.}}(1)
\end{align*}
and 
\begin{align*}
\frac{\log\left(\sum_{j=1}^n \tau_k(j) \right)}{\sqrt{n}} - \frac{\log\left(\sum_{j=1}^n \widehat{\tau}_k(j) \right)}{\sqrt{n}} \overset{\P}{\to} 0.
\end{align*}
So for each sub-optimal arm $k$, the $\tau_k(j)$ and $N_k(t)$ only depend on the rewards for arm $k$.
This establishes the asymptotic independence of the $N_k(t)$ for different sub-optimal arms $k$.
The conclusion (\ref{thm6_2}) then follows from (\ref{thm6_1}) by summing up the contributions of each $N_k(t)$ to the regret $R(t)$.
\halmos
\endproof

\section{Analysis of UCB} \label{ucb}

In this section, we analyze a version of UCB that is designed for iid Gaussian rewards with variance $1$.
We assume that the actual arm rewards are also iid Gaussian with variance $1$, i.e., UCB is operating in a well-specified environment.
(This version of UCB is called UCB1, and it was originally proposed by \cite{auer_etal2002}. It also works for general sub-Gaussian arm reward distributions. With simple tuning, it can handle sub-Gaussian distributions with other variance proxies.)
For modifications and consideration of model mis-specification, see Section \ref{mis}.


Given $\mathcal{F}_t$ (the information collected up to and including time $t$), at time $t+1$ UCB plays the arm with the highest index:
\begin{align*}
    A(t+1) = \argmax_{k \in [K]} \; U_k(N_k(t),t+1),
\end{align*}
where
\begin{align}
U_k(N_k(t),t+1) = \widehat{\mu}_k(N_k(t)) + \sqrt{\frac{2\log(t+1)}{N_k(t)}} \label{klucbef}
\end{align}
and $\widehat{\mu}_k(n) = \frac{1}{n} \sum_{j=1}^n X_k(j)$.
This version of UCB, was introduced as UCB1 in \cite{auer_etal2002}.
As discussed in Chapter 8 of \cite{lattimore_etal2020}, the expected regret for this version of UCB satisfies:
\begin{align}
    \lim_{T \to \infty} \frac{\E[R(T)]}{\log(T)} = \sum_{k \ne k^*} \frac{2}{\Delta_k}. \label{ucb_limit}
\end{align}

\subsection{Strong Law of Large Numbers} \label{ucb_slln}

We first show that the regret of UCB satisfies a SLLN in multi-armed settings.
As with TS, the limit in the SLLN for UCB matches that of expected regret in (\ref{ucb_limit}).
The proof here is adapted from Propositions 7-8 of \cite{cowan_etal2019}.

\begin{theorem} \label{thm3}
Using UCB, for each sub-optimal arm $k$,
\begin{align}
\frac{N_k(T)}{\log(T)} \overset{\text{a.s.}}{\to} \frac{2}{\Delta_k^2}. \label{thm3_1}
\end{align}
Therefore, the regret satisfies the SLLN:
\begin{align*}
\frac{R(T)}{\log(T)} \overset{\text{a.s.}}{\to} \sum_{k \ne k^*} \frac{2}{\Delta_k}. 
\end{align*}
\end{theorem}

\proof{Proof of Theorem \ref{thm3}.}

This proof is an extension and simplification of Propositions 7-8 of \cite{cowan_etal2019}.

We begin with the upper bound part of the proof.
Let $\delta \in (0,\Delta_k/2)$.
For each sub-optimal arm $k$, we have
\begin{align}
    N_k(T) & = 1 + \sum_{t=K}^{T-1} \indic{A(t+1) = k, \; U_k(N_k(t),t+1) \ge \mu_{k^*} - \delta, \; \widehat{\mu}_k(N_k(t)) \le \mu_k + \delta } \label{sum1} \\
    & \quad \;\;\; + \sum_{t=K}^{T-1} \indic{A(t+1) = k, \; U_k(N_k(t),t+1) \ge \mu_{k^*} - \delta, \; \widehat{\mu}_k(N_k(t)) > \mu_k + \delta } \label{sum2} \\ 
    & \quad \;\;\; + \sum_{t=K}^{T-1} \indic{A(t+1) = k, \; U_k(N_k(t),t+1) < \mu_{k^*} - \delta }. \label{sum3}
\end{align}

The first sum is upper bounded via:
\begin{align}
    (\ref{sum1}) & \le \sum_{t=K}^{T-1} \indic{A(t+1) = k, \; (\Delta_k - 2\delta)^2 \le \frac{2\log(t+1)}{N_k(t)} } \label{sum4} \\ 
    & \le \sum_{t=K}^{T-1} \indic{A(t+1) = k, \; N_k(t) \le \frac{2\log(t+1)}{(\Delta_k - 2\delta)^2} } \nonumber \\
    & \le \frac{2\log(t+1)}{(\Delta_k - 2\delta)^2} + 1. \label{sum5}
\end{align}
The bound in (\ref{sum4}) holds due to the events $U_k(N_k(t),t+1) \ge \mu_{k^*} - \delta$ and $\widehat{\mu}_k(N_k(t)) \le \mu_k + \delta$ and the definition of the index in (\ref{klucbef}).

The second sum is upper bounded via:
\begin{align}
    (\ref{sum2}) \le \sum_{t=K}^{T-1} \indic{A(t+1) = k, \; \widehat{\mu}_k(N_k(t)) > \mu_k + \delta }. \label{sum6}
\end{align}
The sum in (\ref{sum6}) can equal $1$ for at most finitely many $t$ by the SLLN for the sample mean and the fact that for each $1$ in the sum, arm $k$ is played an additional time and there is an additional sample that is averaged in the sample mean.

The third sum is upper bounded via:
\begin{align}
    (\ref{sum3}) & \le \sum_{t=K}^{T-1} \indic{A(t+1) = k, \; U_{k^*}(N_{k^*}(t),t+1) \le U_k(N_k(t),t+1) < \mu_{k^*} - \delta } \nonumber \\
    & \le \sum_{t=K}^{T-1} \indic{ U_{k^*}(N_{k^*}(t),t+1) < \mu_{k^*} - \delta }. \label{sum7}
\end{align}
The sum in (\ref{sum7}) can equal $1$ for at most finitely many $t$ by the SLLN for the sample mean and the form of the index in (\ref{klucbef}) with $\log(t)$ increasing.

Putting together (\ref{sum5}), (\ref{sum6}) and (\ref{sum7}) and sending $\delta \downarrow 0$, we have established that almost surely for each sub-optimal arm $k$,
\begin{align}
    \limsup_{T \to \infty} \frac{N_k(T)}{\log(T)} \le \frac{2}{\Delta_k^2}. \label{suboptimallimsup}
\end{align}
Therefore, for the optimal arm $k^*$, almost surely,
\begin{align}
    \lim_{T \to \infty} \frac{N_{k^*}(T)}{T} = 1, \label{optimalconvergence}
\end{align}
which then implies by the form of the index in (\ref{klucbef}) and the SLLN for the sample mean that almost surely,
\begin{align*}
    \lim_{t \to \infty} U_{k^*}(N_{k^*}(t),t+1) = \mu_{k^*}. 
\end{align*}
This also implies that almost surely, all sub-optimal arms are played infinitely many times, due to the term $\log(t)$ growing without bound in the index (\ref{klucbef}).

We now develop the lower bound parts of the proof.
For all sub-optimal arms $k$,
\begin{align}
    U_{k^*}(N_{k^*}(\mathcal{T}_{k^*}(j)-1),\mathcal{T}_{k^*}(j)) > U_k(N_k(\mathcal{T}_{k^*}(j)-1),\mathcal{T}_{k^*}(j)). \label{taum}
\end{align}
We have for sufficiently large $j$, almost surely,
\begin{align}
    \max_{t \in [\mathcal{T}_{k^*}(j),\mathcal{T}_{k^*}(j+1)]} \frac{\log(t)}{N_k(t)} & \le \frac{\log(\mathcal{T}_{k^*}(j+1))}{N_k(\mathcal{T}_{k^*}(j)-1)} \nonumber \\
    & = \frac{\log(\mathcal{T}_{k^*}(j+1))}{\log(\mathcal{T}_{k^*}(j))} \frac{\log(\mathcal{T}_{k^*}(j))}{N_k(\mathcal{T}_{k^*}(j)-1)} \nonumber \\
    & \le (1+\delta) \frac{\log(\mathcal{T}_{k^*}(j))}{N_k(\mathcal{T}_{k^*}(j)-1)} \label{loglog} \\ 
    & \le (1+\delta) \frac{1}{2} (U_k(N_k(\mathcal{T}_{k^*}(j)-1),\mathcal{T}_{k^*}(j)) - \mu_k + \delta)^2 \label{d1} \\
    & \le (1+\delta) \frac{1}{2} (U_{k^*}(N_{k^*}(\mathcal{T}_{k^*}(j)-1),\mathcal{T}_{k^*}(j)) - \mu_k + \delta)^2 \label{d2} \\
    & \le (1+\delta) \frac{1}{2} (\Delta_k + 2\delta)^2. \label{d3}
\end{align}
Note that (\ref{loglog}) is due to (\ref{optimalconvergence}), (\ref{d1}) is due to the SLLN for the sample mean of the sub-optimal arm $k$ and the form of the index in (\ref{klucbef}), (\ref{d2}) is due to (\ref{taum}), and (\ref{d3}) is due to the SLLN for the sample mean of the optimal arm $k^*$.
Therefore, sending $j \to \infty$ and $\delta \downarrow 0$, almost surely,
\begin{align*}
    \liminf_{T \to \infty} \frac{N_k(T)}{\log(T)} \ge \frac{2}{\Delta_k^2}, 
\end{align*}
which together with (\ref{suboptimallimsup}), establishes (\ref{thm3_1}).
\halmos
\endproof

\subsection{Central Limit Theorem} \label{ucb_clt}

We now show that the regret of UCB satisfies a CLT in two-armed settings.
The (mean) centering in the CLT matches the asymptotics of the SLLN in Theorem \ref{thm3} (and also that of expected regret in (\ref{ucb_limit})).
To prove the CLT, we directly analyze the times $\mathcal{T}_k(j)$ during which the sub-optimal arm $k$ is played.
We will see that these times are determined by what is essentially a perturbed random walk with positive drift.
Again, it turns out that the variability in the CLT is purely due to the variability of the rewards of the sub-optimal arm.
And this is reasonable in light of the fact that the optimal arm is played much more than the sub-optimal arm, and so its sample mean is much more concentrated around its true mean.

\begin{theorem} \label{thm4}
In two-armed bandit environments, the regret of UCB satisfies the CLT:
\begin{align}
\frac{R(T) - \frac{2}{\Delta} \log(T)}{\frac{2}{\Delta}\sqrt{2\log(T)}} \Rightarrow N(0,1). \label{thm4_0}
\end{align}
\end{theorem}

\proof{Proof of Theorem \ref{thm4}.}
Without loss of generality, let arm $2$ be the sub-optimal arm.
We first establish a few preliminaries.
\begin{align}
& \mathcal{T}_2(j+1) \nonumber \\
& = \inf\left\{ t \; : \; t \in \mathbb{Z}_+, \; t > \mathcal{T}_2(j), \; \widehat{\mu}_2(j) + \sqrt{\frac{2 \log t}{j}} > \widehat{\mu}_1(N_1(t-1)) + \sqrt{\frac{2 \log t}{N_1(t-1)}} \right\} \nonumber \\
& = \inf\left\{ t \; : \; t \in \mathbb{Z}_+, \; t > \mathcal{T}_2(j), \; t > \exp\left(\frac{j}{2} \left( \Delta_2 - \Bigl(\widehat{\mu}_2(j) - \mu_2\Bigr) + \Bigl(\widehat{\mu}_1(N_1(t-1)) - \mu_1\Bigr) + \sqrt{\frac{2 \log t}{N_1(t-1)}} \right)^2\right) \right\} \nonumber \\
& = 1 + \left\lfloor \mathcal{T}_2(j) \vee \exp(S_2(j)) \right\rfloor, \label{ucb3}
\end{align}
where we define
\begin{align*}
S_2(j) = \frac{j}{2} \left( \Delta_2 - \Bigl(\widehat{\mu}_2(j) - \mu_2 \Bigr) + \Bigl(\widehat{\mu}_1(N_1(\mathcal{T}_2(j+1)-1)) - \mu_1\Bigr) + \sqrt{\frac{2 \log(\mathcal{T}_2(j+1))}{N_1(\mathcal{T}_2(j+1)-1)}} \right)^2. 
\end{align*}
Expanding the square, 
\begin{align}
    S_2(j) = \Delta_2 \sum_{i=1}^j \left( Y_2(i) + \frac{\Delta_2}{2} \right) + E_2(j), \label{ucb4}
\end{align}
where $Y_2(i) = -(X_2(i) - \mu_2)$ is an independent sequence of $N(0,1)$ random variables.
And the sequence of random variables $E_2(j)$ satisfies: 
\begin{align}
    E_2(j+1) - E_2(j) \overset{\text{a.s.}}{\to} 0, \label{ucb5}
\end{align}
which follows from the LIL and the conclusions from Theorem \ref{thm3}:
\begin{align*}
& \frac{N_2(t)}{\log(t)} \overset{\text{a.s.}}{\to} \frac{2}{\Delta_2^2} \\
& \frac{N_1(t)}{t} \overset{\text{a.s.}}{\to} 1.
\end{align*}

The main task is to show that
\begin{align}
\frac{\log\left(\mathcal{T}_2(j)\right) - j\frac{\Delta_2^2}{2}}{\Delta_2 \sqrt{j}} \Rightarrow N(0,1). \label{ucb_main}
\end{align}
This is established in Lemma \ref{lem5}.
As in the proof of Theorem \ref{thm2}, let $x \in \mathbb{R}$, and define $h(t) = \floor{x \frac{2}{\Delta_2^{2}} \sqrt{2t} + \frac{2}{\Delta_2^2} t}$.
Then,
\begin{align}
P\left( \frac{\log\left(\mathcal{T}_2(h(t))\right) - h(t)\frac{\Delta_2^2}{2}}{\Delta_2\sqrt{h(t)}} > -x \right) & \sim P\left( \frac{\log\left(\mathcal{T}_2(h(t))\right) - h(t)\frac{\Delta_2^2}{2}}{\Delta_2\sqrt{h(t)}} > \frac{t - h(t)\frac{\Delta_2^2}{2}}{\Delta_2\sqrt{h(t)}} \right) \nonumber \\ 
& = P\left( \mathcal{T}_2(h(t)) > e^t \right) \nonumber \\
& = P\left( N_2(e^t) \le h(t) \right) \nonumber \\
& = P\left(\frac{N_2(e^t) - t \frac{2}{\Delta_2^2}}{\frac{2}{\Delta_2^2} \sqrt{2t}} \le x \right). \label{conclusion_0}
\end{align}
Using this together with (\ref{ucb_main}), (\ref{thm4_0}) is established.
\halmos
\endproof

\begin{lemma} \label{lem5}
Using UCB in two-armed bandit environments in which arm $2$ is sub-optimal,
\begin{align}
\frac{\log(\mathcal{T}_2(j)) - j\frac{\Delta_2^2}{2}}{\sqrt{j}} \Rightarrow N(0,1). \label{lem5_0}
\end{align}
\end{lemma}

\proof{Proof of Lemma \ref{lem5}.}
We first establish some preliminary facts.
For each positive integer index $l$, define
\begin{align}
j_l^* = \inf\{ j > j_{l-1}^* : \exp(S_2(j)) \ge \mathcal{T}_2(j) \}, \label{ucb6}
\end{align}
so that from (\ref{ucb3}), the $j_l^*$ are precisely those instances $j$ satisfying
\begin{align*}
\mathcal{T}_2(j+1) = 1 + \left\lfloor \exp(S_2(j)) \right\rfloor.
\end{align*}
Note that
\begin{align}
& j_{l+1}^* - j_l^* \nonumber \\
& = \inf\left\{ i \ge 1 : \exp(S_2(j_l^* + i)) > 1 + \lfloor\exp(S_2(j_l^*))\rfloor + i \right\} \nonumber \\
& = \inf\left\{ i \ge 1 : S_2(j_l^* + i) > S_2(j_l^*) + \log(1 + \lfloor\exp(S_2(j_l^*))\rfloor + i) - S_2(j_l^*) \right\} \nonumber \\
& = \inf\left\{ i \ge 1 : \Delta_2 \sum_{m=1}^i \left(Y_2(j_l^* + m) + \frac{\Delta_2}{2}\right) > E_2(j_l^*) - E_2(j_l^* + i) + \log(1 + \lfloor\exp(S_2(j_l^*))\rfloor + i) - S_2(j_l^*) \right\}, \label{ucb8}
\end{align}
where in (\ref{ucb8}), we have used the definition of $E_2(j)$ as a component of $S_2(j)$, as expressed in (\ref{ucb4}).
Also define
\begin{align*}
M_2(l) & = \inf\left\{ i \ge 1 : \Delta_2 \sum_{m=1}^i \left(Y_2(j_l^* + m) + \frac{\Delta_2}{2}\right) > i \frac{\Delta_2^2}{4} \right\}. 
\end{align*}
Examining (\ref{ucb8}), using (\ref{ucb5}) and the fact that $\log(i + 1) - \log(i) \overset{\text{a.s.}}{\to} 0$, we have
\begin{align*}
\abs{E_2(j_l^*) - E_2(j_l^* + i) + \log(1 + \lfloor\exp(S_2(j_l^*))\rfloor + i) - S_2(j_l^*)} \le i \frac{\Delta_2^2}{4},
\end{align*}
almost surely, for sufficiently large $l$ and all $i \ge 1$, and thus also,
\begin{align}
j_{l+1}^* - j_l^* \le M_2(l). \label{ucb10}
\end{align}

Proceeding with the main parts of the proof, there are two cases to consider.
In the first case, if $j = j_l^*$ for some $j_l^*$, then
\begin{align}
\log(\mathcal{T}_2(j)) \vee S_2(j) = S_2(j). \label{ucb12}
\end{align}
In the second case, if $j$ is such that $j_l^* < j < j_{l+1}^*$ for some $j_l^*$ and $j_{l+1}^*$, then 
\begin{align*}
\log(\mathcal{T}_2(j)) \vee S_2(j) = \log(\mathcal{T}_2(j)).
\end{align*}
And almost surely, for sufficiently large $j$ (and hence, sufficiently large $j_l^*$), we have
\begin{align}
0 & \le \log(\mathcal{T}_2(j)) - S_2(j) \nonumber \\
& \le \left(\log\Bigl(\mathcal{T}_2(j_l^* + 1) + M_2(l)\Bigr) - S_2(j_l^*)\right) - \Bigl(S_2(j) - S_2(j_l^*)\Bigr) \label{ucb13} \\
& = \left(\log\Bigl(1 + \left\lfloor \exp{S_2(j_l^*)} \right\rfloor + M_2(l)\Bigr) - S_2(j_l^*)\right) - \Bigl(S_2(j) - S_2(j_l^*)\Bigr) \label{ucb14} \\
& \le \left(\log\Bigl(1 + \left\lfloor \exp(S_2(j_l^*)) \right\rfloor + M_2(l)\Bigr) - S_2(j_l^*)\right) - \inf_{1 \le i \le M_2(l)}\left\{\Delta_2 \sum_{m=1}^{i} \left(Y_2(j_l^* + m) + \frac{\Delta_2}{2}\right) \right\} + M_2(l) \frac{\Delta_2^2}{4}. \label{ucb15}
\end{align}
Note that (\ref{ucb13}) follows from (\ref{ucb10}), which is an upper bound on $j_{l+1}^* - j_l^*$, as well as the fact that for $j$ such that $j_l^* < j < j_{l+1}^*$,
\begin{align*}
\mathcal{T}_2(j) = \mathcal{T}_2(j-1) + 1.
\end{align*}
This fact is true because of the identity (\ref{ucb3}), together with the definition of $j_l^*$ in (\ref{ucb6}), which implies that for $j$ such that $j_l^* < j < j_{l+1}^*$,
\begin{align*}
\exp(S_2(j)) < \mathcal{T}_2(j).
\end{align*}
Also, (\ref{ucb14}) follows from the definition of $j_l^*$ as satisfying:
\begin{align*}
\mathcal{T}_2(j_l^*+1) = 1 + \left\lfloor \exp(S_2(j_l^*)) \right\rfloor.
\end{align*}
And (\ref{ucb15}) follows from the relation in (\ref{ucb4}), so that
\begin{align*}
S_2(j) - S_2(j_l^*) = \Delta_2 \sum_{m=1}^{j-j_l^*} \left(Y_2(j_l^* + m) + \frac{\Delta_2}{2}\right) + E_2(j) - E_2(j_l^*),
\end{align*}
along with the fact that almost surely, for sufficiently large $j_l$, 
\begin{align*}
\abs{E_2(j) - E_2(j_l^*)} \le (j - j_l^*) \frac{\Delta_2^2}{4} \le M_2(l) \frac{\Delta_2^2}{4}.
\end{align*}
Using (\ref{ucb15}), and the fact that for each $l$, the $M_2(l)$ and $Y_2(j_l^* + m)$, $1 \le m \le M_2(l)$ are iid random variables, we have
\begin{align}
\frac{\log(\mathcal{T}_2(j))}{\sqrt{j}} - \frac{S_2(j)}{\sqrt{j}} \overset{\P}{\to} 0 \label{ucb16}
\end{align}
as $j \to \infty$ along sequences of $j$ such that $j_l^* < j < j_{l+1}^*$ for some $j_l^*$ and $j_{l+1}^*$.
Putting together (\ref{ucb16}) and (\ref{ucb12}) (for the cases that $j = j_l^*$ for some $j_l^*$), we have shown that
\begin{align}
\frac{\log(\mathcal{T}_2(j)) \vee S_2(j)}{\sqrt{j}} - \frac{S_2(j)}{\sqrt{j}} \overset{\P}{\to} 0 \label{ucb11}
\end{align}
as $j \to \infty$ (without restrictions on $j$).
Also, from (\ref{ucb4}), we have
\begin{align}
\frac{S_2(j)}{\sqrt{j}} = \frac{\Delta_2 \sum_{i=1}^j \left( Y_2(i) + \frac{\Delta_2}{2} \right) + E_2(j)}{\sqrt{j}}, \label{ucb0}
\end{align}
where $E_2(j)/\sqrt{j} \overset{\text{a.s.}}{\to} 0$.
Then, (\ref{lem5_0}) is established using (\ref{ucb11}) and (\ref{ucb0}) together with
\begin{align*}
\frac{\log(\mathcal{T}_2(j+1))}{\sqrt{j}} - \frac{\log(\mathcal{T}_2(j)) \vee S_2(j)}{\sqrt{j}} \overset{\text{a.s.}}{\to} 0,
\end{align*}
which is obtained from (\ref{ucb3}).
\halmos
\endproof

\subsection{Extension to Multiple Arms} \label{ucb_multi}

In this section, we extend the CLT for the regret of UCB in two-armed settings (Theorem \ref{thm4}) to multi-armed settings.
The key to the extension is the fact that once the UCB index of a sub-optimal arm exceeds that of the optimal arm, it is guaranteed that the particular sub-optimal arm will be played relatively soon (if not immediately).
Although there could simultaneously be other sub-optimal arms with indices higher than that of the optimal arm, these other arms cannot delay the play of the particular sub-optimal arm by too long.
So, effectively, each sub-optimal arm only competes with the optimal arm to be played, and the analysis in multi-armed settings reduces to that in the two-armed setting.
As is the case for TS, $N_k(T)$ for each sub-optimal arm $k$ depends only on the rewards received for that arm.
So again, the $N_k(T)$ of different sub-optimal arms $k$ are independent and contribute additively to the overall CLT variance.

\begin{theorem} \label{thm7}
Using UCB, for each sub-optimal arm $k$,
\begin{align}
\frac{N_k(T) - \frac{2}{\Delta_k^2} \log(T)}{\frac{2}{\Delta_k^2}\sqrt{2\log(T)}} \Rightarrow N(0,1). \label{thm7_1}
\end{align}
Furthermore, for different sub-optimal arms $k$, the $N_k(T)$ are asymptotically independent.
Therefore, the regret satisfies the CLT:
\begin{align}
\frac{R(T) - \sum_{k \ne k^*} \frac{2}{\Delta_k} \log(T)}{\sqrt{\sum_{k \ne k^*} \frac{8}{\Delta_k^2} \log(T)}} \Rightarrow N(0,1). \label{thm7_2}
\end{align}
\end{theorem}

\proof{Proof of Theorem \ref{thm7}.}
For any sub-optimal arm $k$, let $\epsilon = \exp(\frac{\Delta_k^2}{8}) - 1$.
Define
\begin{align}
\mathcal{T}^{-}_k(j+1) & = \inf\left\{ t \; : \; t \in \mathbb{Z}_+, \; t > \mathcal{T}^{-}_k(j), \; U_k(j,t+1) > U_{k^*}(N_{k^*}(t),t+1) \right\} \label{ucb_13} \\
\mathcal{T}^{+}_k(j+1) & = \inf\left\{ (1+\epsilon)t \; : \; t \in \mathbb{R}_+, \; t \ge \mathcal{T}^{+}_k(j), \; U_k(j,t+1) \ge U_{k^*}(N_{k^*}(\lfloor t \rfloor),t+1) \right\}. \label{ucb_14}
\end{align}
Recall that
\begin{align}
\mathcal{T}_k(j+1) = \inf\left\{ t \; : \; t \in \mathbb{Z}_+, \; t > \mathcal{T}_k(j), \; U_k(j,t+1) > \max_{k' \ne k} \; U_{k'}(N_{k'}(t),t+1) \right\}. \nonumber
\end{align}
Almost surely, for $j$ sufficiently large,
\begin{align}
    \mathcal{T}^{-}_k(j) \le \mathcal{T}_k(j) \le \mathcal{T}^{+}_k(j). \label{ucb_15}
\end{align}
It is straightforward to see that the lower bound on $\mathcal{T}_k(j)$ in (\ref{ucb_15}) holds.
The upper bound holds by the following argument.
Each time the UCB index for arm $k$ exceeds that of arm $k^*$, it is guaranteed that arm $k$ will be played before the next time that arm $k^*$ is played.
The only possible delay to arm $k$ being played immediately upon its UCB index exceeding that of arm $k^*$ is if there are also other sub-optimal arms with their UCB indices exceeding that of arm $k^*$.
These other sub-optimal arms could compete with arm $k$ to be played, thus potentially delaying plays of arm $k$.
However, from Theorem \ref{thm3}, for each sub-optimal arm $k' \ne k$, $N_{k'}(t) \le \frac{4}{\Delta_{k'}^2} \log(t)$ almost surely for $t$ sufficiently large.
Moreover, $\epsilon t > \sum_{k' \ne k, k^*} \frac{4}{\Delta_{k'}^2} \log(t)$ for sufficiently large $t$.
So for sufficiently large $t$, the delay cannot be longer than $\epsilon t$.
Accordingly, the times $\mathcal{T}^{+}_k(j)$ (in (\ref{ucb_14})) are delayed by a $1+\epsilon$ multiplicative factor compared to the times $\mathcal{T}^{-}_k(j)$ (in (\ref{ucb_13})).
Thus, the upper bound on $\mathcal{T}_k(j)$ in (\ref{ucb_15}) must hold for sufficiently large $j$.

Applying the analysis from Lemma \ref{lem5} to $\mathcal{T}^{-}(j)$, we have
\begin{align*}
    \frac{\log(\mathcal{T}^{-}_k(j)) - j \frac{\Delta_k^2}{2}}{\sqrt{j}} \Rightarrow N(0,1).
\end{align*}
Together with Lemma \ref{lem6} and (\ref{ucb_15}), we obtain
\begin{align*}
    \frac{\log(\mathcal{T}_k(j)) - j \frac{\Delta_k^2}{2}}{\sqrt{j}} \Rightarrow N(0,1).
\end{align*}
Then, (\ref{thm7_1}) is established using arguments leading up to (\ref{conclusion_0}) in the proof of Theorem \ref{thm4}.
From the proofs of Lemmas \ref{lem5} and \ref{lem6},
\begin{align*}
    & \frac{\log(\mathcal{T}^{-}_k(j))}{\sqrt{j}} - \frac{S^{-}_k(j)}{\sqrt{j}} \overset{\P}{\to} 0 \\
    & \frac{\log(\mathcal{T}^{+}_k(j))}{\sqrt{j}} - \frac{S^{+}_k(j)}{\sqrt{j}} \overset{\P}{\to} 0 
\end{align*}
where
\begin{align*}
    S^{-}_k(j) & = \Delta_k \sum_{i=1}^j \left( Y_k(i) + \frac{\Delta_k}{2} \right) + E^{-}_k(j) \\
    S^{+}_k(j) & = \Delta_k \sum_{i=1}^j \left( Y_k(i) + \frac{\Delta_k}{2} \right) + E^{+}_k(j),
\end{align*}
with $Y_k(i) = -(X_k(i) - \mu_k)$, and $(E^{-}_k(j) + E^{+}_k(j))/\sqrt{j} \overset{\text{a.s.}}{\to} 0$.
This establishes the asymptotic independence of $N_k(t)$ for different sub-optimal arms $k$.
Then, (\ref{thm7_2}) follows from summing up the contributions of each $N_k(t)$ to the regret $R(t)$.
\halmos
\endproof

\begin{lemma} \label{lem6}
Using UCB, for each sub-optimal arm $k$, with $\mathcal{T}^{+}_k(j)$ as defined in (\ref{ucb_14}),
\begin{align}
\frac{\log(\mathcal{T}^{+}_k(j)) - j \frac{\Delta_k^2}{2}}{\sqrt{j}} \Rightarrow N(0,1). \label{lem6_1}
\end{align}
\end{lemma}

\proof{Proof of Lemma \ref{lem6}.}
Note that
\begin{align}
    & \mathcal{T}^{+}_k(j+1) \nonumber \\
    & = \inf\Biggl\{ (1+\epsilon)t \; : \; t \in \mathbb{R}_+, \; t \ge \mathcal{T}^{+}_k(j), \; t \ge \exp\Biggl(\frac{j}{2} \biggl( \Delta_k - \Bigl(\widehat{\mu}_k(j) - \mu_k\Bigr) + \nonumber \\
    & \qquad \qquad \qquad \qquad \qquad \qquad \qquad \qquad \qquad \qquad \quad \Bigl(\widehat{\mu}_{k^*}(N_{k^*}(\lfloor t \rfloor - 1)) - \mu_{k^*}\Bigr) + \sqrt{\frac{2 \log(t)}{N_{k^*}(\lfloor t \rfloor - 1)}} \biggr)^2\Biggr) \Biggr\} \nonumber \\
    & = (1+\epsilon)\Bigl( \mathcal{T}^{+}_k(j) \vee \exp(S^{+}_k(j)) \Bigr), \label{ucb_00}
\end{align}
where
\begin{align*}
    S^{+}_k(j) = \frac{j}{2} \left( \Delta_k - \Bigl(\widehat{\mu}_k(j) - \mu_k\Bigr) + \Bigl(\widehat{\mu}_{k^*}(N_{k^*}(\lfloor \mathcal{T}^{+}_k(j+1)/(1+\epsilon) \rfloor - 1)) - \mu_{k^*}\Bigr) + \sqrt{\frac{2 \log(\frac{\mathcal{T}^{+}_k(j+1)}{1+\epsilon})}{N_{k^*}(\lfloor \frac{\mathcal{T}^{+}_k(j+1)}{1+\epsilon} \rfloor - 1)}} \right)^2. 
\end{align*}
Expanding the square, 
\begin{align}
    S^{+}_k(j) = \Delta_k \sum_{i=1}^j \left( Y_k(i) + \frac{\Delta_k}{2} \right) + E^{+}_k(j), \label{ucb_1}
\end{align}
where $Y_k(i) = -(X_k(i) - \mu_k)$ is an independent sequence of $N(0,1)$ random variables.
And the sequence of random variables $E^{+}_k(j)$ satisfies: 
\begin{align}
    E^{+}_k(j+1) - E^{+}_k(j) \overset{\text{a.s.}}{\to} 0, \label{ucb_2}
\end{align}
which follows from the LIL and the conclusions from Theorem \ref{thm3}.

For each positive integer index $l$, define
\begin{align}
j_l^* = \inf\{ j > j_{l-1}^* : \exp(S^{+}_k(j)) \ge \mathcal{T}^{+}_k(j) \}, \label{ucb_000}
\end{align}
so that from (\ref{ucb_00}), the $j_l^*$ are precisely those instances $j$ satisfying
\begin{align*}
\mathcal{T}^{+}_k(j+1) = (1+\epsilon)\exp(S^{+}_k(j)).
\end{align*}
Note that
\begin{align}
j_{l+1}^* - j_l^* & = \inf\left\{ i \ge 1 : \exp(S^{+}_k(j_l^* + i)) \ge (1+\epsilon)^i \exp(S^{+}_k(j_l^*)) \right\} \nonumber \\
& = \inf\left\{ i \ge 1 : S^{+}_k(j_l^* + i) > S^{+}_k(j_l^*) + i\log(1+\epsilon) \right\} \nonumber \\
& = \inf\left\{ i \ge 1 : \Delta_k \sum_{m=1}^i \left(Y_k(j_l^* + m) + \frac{\Delta_k}{2}\right) > E^{+}_k(j_l^*) - E^{+}_k(j_l^* + i) + i \log(1+\epsilon) \right\}, \label{ucb_3}
\end{align}
where in (\ref{ucb_3}), we have used the definition of $E^{+}_k(j)$ as a component of $S^{+}_k(j)$, as expressed in (\ref{ucb_1}).
Also define
\begin{align*}
M_k(l) & = \inf\left\{ i \ge 1 : \Delta_k \sum_{m=1}^i \left(Y_k(j_l^* + m) + \frac{\Delta_k}{2}\right) > i \frac{\Delta_k^2}{4} \right\}. 
\end{align*}
Using (\ref{ucb_2}), we have
\begin{align*}
\abs{E^{+}_k(j_l^*) - E^{+}_k(j_l^* + i) + i \log(1+\epsilon)} \le i \frac{\Delta_k^2}{4},
\end{align*}
almost surely, for sufficiently large $l$ and all $i \ge 1$, and thus also,
\begin{align}
j_{l+1}^* - j_l^* \le M_k(l). \label{ucb_5}
\end{align}

Proceeding with the main parts of the proof, there are two cases to consider.
In the first case, if $j = j_l^*$ for some $j_l^*$, then
\begin{align}
\log(\mathcal{T}^{+}_k(j)) \vee S^{+}_k(j) = S^{+}_k(j). \label{ucb_6}
\end{align}
In the second case, if $j$ is such that $j_l^* < j < j_{l+1}^*$ for some $j_l^*$ and $j_{l+1}^*$, then 
\begin{align*}
\log(\mathcal{T}^{+}_k(j)) \vee S^{+}_k(j) = \log(\mathcal{T}^{+}_k(j)).
\end{align*}
And almost surely, for sufficiently large $j$ (and hence, sufficiently large $j_l^*$), we have
\begin{align}
0 & \le \log(\mathcal{T}^{+}_k(j)) - S^{+}_k(j) \nonumber \\
& \le \left(\log\Bigl((1+\epsilon)^{M_k(l)} \mathcal{T}^{+}_k(j_l^* + 1)\Bigr) - S^{+}_k(j_l^*)\right) - \Bigl(S^{+}_k(j) - S^{+}_k(j_l^*)\Bigr) \label{ucb_7} \\
& = (M_k(l) + 1)\log(1+\epsilon) - \Bigl(S^{+}_k(j) - S^{+}_k(j_l^*)\Bigr) \label{ucb_8} \\
& \le (M_k(l) + 1)\log(1+\epsilon) - \inf_{1 \le i \le M_k(l)}\left\{\Delta_k \sum_{m=1}^{i} \left(Y_k(j_l^* + m) + \frac{\Delta_k}{2}\right) \right\} + M_k(l) \frac{\Delta_k^2}{4}. \label{ucb_9}
\end{align}
Note that (\ref{ucb_7}) follows from (\ref{ucb_5}), which is an upper bound on $j_{l+1}^* - j_l^*$, as well as the fact that for $j$ such that $j_l^* < j < j_{l+1}^*$,
\begin{align*}
\mathcal{T}^{+}_k(j) = (1+\epsilon) \mathcal{T}^{+}_k(j-1).
\end{align*}
This fact is true because of the identity (\ref{ucb_00}), together with the definition of $j_l^*$ in (\ref{ucb_000}), which implies that for $j$ such that $j_l^* < j < j_{l+1}^*$,
\begin{align*}
\exp(S^{+}_k(j)) < \mathcal{T}^{+}_k(j).
\end{align*}
Also, (\ref{ucb_8}) follows from the definition of $j_l^*$ as satisfying:
\begin{align*}
\mathcal{T}^{+}_k(j_l^*+1) = (1+\epsilon) \exp(S^{+}_k(j_l^*)).
\end{align*}
And (\ref{ucb_9}) follows from the relation in (\ref{ucb_1}), so that
\begin{align*}
S^{+}_k(j) - S^{+}_k(j_l^*) = \Delta_k \sum_{m=1}^{j-j_l^*} \left(Y_k(j_l^* + m) + \frac{\Delta_k}{2}\right) + E^{+}_k(j) - E^{+}_k(j_l^*),
\end{align*}
along with the fact that almost surely, for sufficiently large $j_l$, 
\begin{align*}
\abs{E^{+}_k(j) - E^{+}_k(j_l^*)} \le (j - j_l^*) \frac{\Delta_k^2}{4} \le M_k(l) \frac{\Delta_k^2}{4}.
\end{align*}
Using (\ref{ucb_9}), and the fact that for each $l$, the $M_k(l)$ and $Y_k(j_l^* + m)$, $1 \le m \le M_k(l)$ are iid random variables, we have
\begin{align}
\frac{\log(\mathcal{T}^{+}_k(j))}{\sqrt{j}} - \frac{S^{+}_k(j)}{\sqrt{j}} \overset{\P}{\to} 0 \label{ucb_10}
\end{align}
as $j \to \infty$ along sequences of $j$ such that $j_l^* < j < j_{l+1}^*$ for some $j_l^*$ and $j_{l+1}^*$.
Putting together (\ref{ucb_10}) and (\ref{ucb_6}) (for the cases that $j = j_l^*$ for some $j_l^*$), we have shown that
\begin{align}
\frac{\log(\mathcal{T}^{+}_k(j)) \vee S^{+}_k(j)}{\sqrt{j}} - \frac{S^{+}_k(j)}{\sqrt{j}} \overset{\P}{\to} 0 \label{ucb_11}
\end{align}
as $j \to \infty$ (without restrictions on $j$).
Also, from (\ref{ucb_1}), we have
\begin{align}
\frac{S^{+}_k(j)}{\sqrt{j}} = \frac{\Delta_k \sum_{i=1}^j \left( Y_k(i) + \frac{\Delta_k}{2} \right) + E^{+}_k(j)}{\sqrt{j}}, \label{ucb_12}
\end{align}
where $\frac{E^{+}_k(j)}{\sqrt{j}} \overset{\text{a.s.}}{\to} 0$.
Then, (\ref{lem6_1}) is established using (\ref{ucb_11}) and (\ref{ucb_12}) together with (\ref{ucb_00}).
\halmos
\endproof

\section{Modifications and Model Mis-specification} \label{mis}

In this section, we develop SLLN's in Proposition \ref{prop1} and CLT's in Proposition \ref{prop2} for the regret of TS and UCB tuned for Gaussian rewards with variance $\sigma^2$.
For the SLLN's, the rewards for each arm $k$ can have an arbitrarily distribution with finite mean $\mu_k$.
For the CLT's, there is the additional requirement of a finite variance $\sigma_k^2$ for each arm $k$.
The proofs of Propositions \ref{prop1} and \ref{prop2} are straightforward modifications of those of Theorems \ref{thm5}, \ref{thm6}, \ref{thm3} and \ref{thm7}, and are thus omitted.

In the SLLN's in Proposition \ref{prop1}, we see that designing for Gaussian rewards with larger variance $\sigma^2$ increases the amount of regret accumulated in the long run.
We see a similar effect in the CLT's in Proposition \ref{prop2}, along with an increase in the CLT variance.
Additionally, increasing the actual reward variances $\sigma_k^2$ also increases the CLT variance.
Nevertheless, the increases in the regret and regret variance in the SLLN's and CLT's due to increasing $\sigma^2$ are counter-balanced by lighter regret distribution tails, as we know from \cite{fan_etal2021b}.
For example, if the rewards are Gaussian with common variance $\sigma_0^2$ for all arms, then using UCB designed for variance $\sigma^2$ Gaussian rewards will yield a regret distribution with tail exponent $-\sigma^2/\sigma_0^2$.
Specifically, $\log \P(R(T) > x) / \log(x) \to -\sigma^2/\sigma_0^2$ uniformly for $x > \log^{1+\epsilon}(T)$ (for arbitrarily small, fixed $\epsilon > 0$) as $T \to \infty$.
(See Corollary 1 and also the more general results in Section 5 of \cite{fan_etal2021b}.)

The SLLN's and CLT's here are quite robust to model mis-specification, and the limits change in a continuous manner in response to changes in the algorithm design and/or reward distributions.
This is in contrast to expected regret, which can be highly sensitive to such changes.
As can be seen via the tail approximations for the regret distribution developed in \cite{fan_etal2021b}, when the bandit environment is just slightly mis-specified relative to the algorithm design, the expected regret can change from scaling as $\log(T)$ to scaling as $T^a$ for some $0 < a < 1$.

\begin{proposition} \label{prop1}
Using either TS or UCB designed for Gaussian rewards with variance $\sigma^2$, for each sub-optimal arm $k$,
\begin{align}
\frac{N_k(T)}{\log(T)} \overset{\text{a.s.}}{\to} \frac{2\sigma^2}{\Delta_k^2}. \label{prop1_1}
\end{align}
Therefore, the regret satisfies the SLLN:
\begin{align}
\frac{R(T)}{\log(T)} \overset{\text{a.s.}}{\to} \sum_{k \ne k^*} \frac{2\sigma^2}{\Delta_k}. \label{prop1_2}
\end{align}
\end{proposition}

\begin{proposition} \label{prop2}
Suppose the rewards for arm $k$ have variance $\sigma_k^2$.
Using either TS or UCB designed for Gaussian rewards with variance $\sigma^2$, for each sub-optimal arm $k$,
\begin{align}
\frac{N_k(T) - \frac{2\sigma^2}{\Delta_k^2} \log(T)}{\frac{2\sigma \sigma_k}{\Delta_k^2}\sqrt{2\log(T)}} \Rightarrow N(0,1). \label{prop2_1}
\end{align}
Furthermore, for different sub-optimal arms $k$, the $N_k(T)$ are asymptotically independent.
Therefore, the regret satisfies the CLT:
\begin{align}
\frac{R(T) - \sum_{k \ne k^*} \frac{2\sigma^2}{\Delta_k} \log(T)}{\sqrt{\sum_{k \ne k^*} \frac{8 \sigma^2 \sigma_k^2}{\Delta_k^2} \log(T)}} \Rightarrow N(0,1). \label{prop2_2}
\end{align}
\end{proposition}

\section{Numerical Simulations} \label{numerical}

\begin{figure}
     \centering
     \begin{subfigure}[b]{0.4\textwidth}
         \centering
         \includegraphics[width=\textwidth]{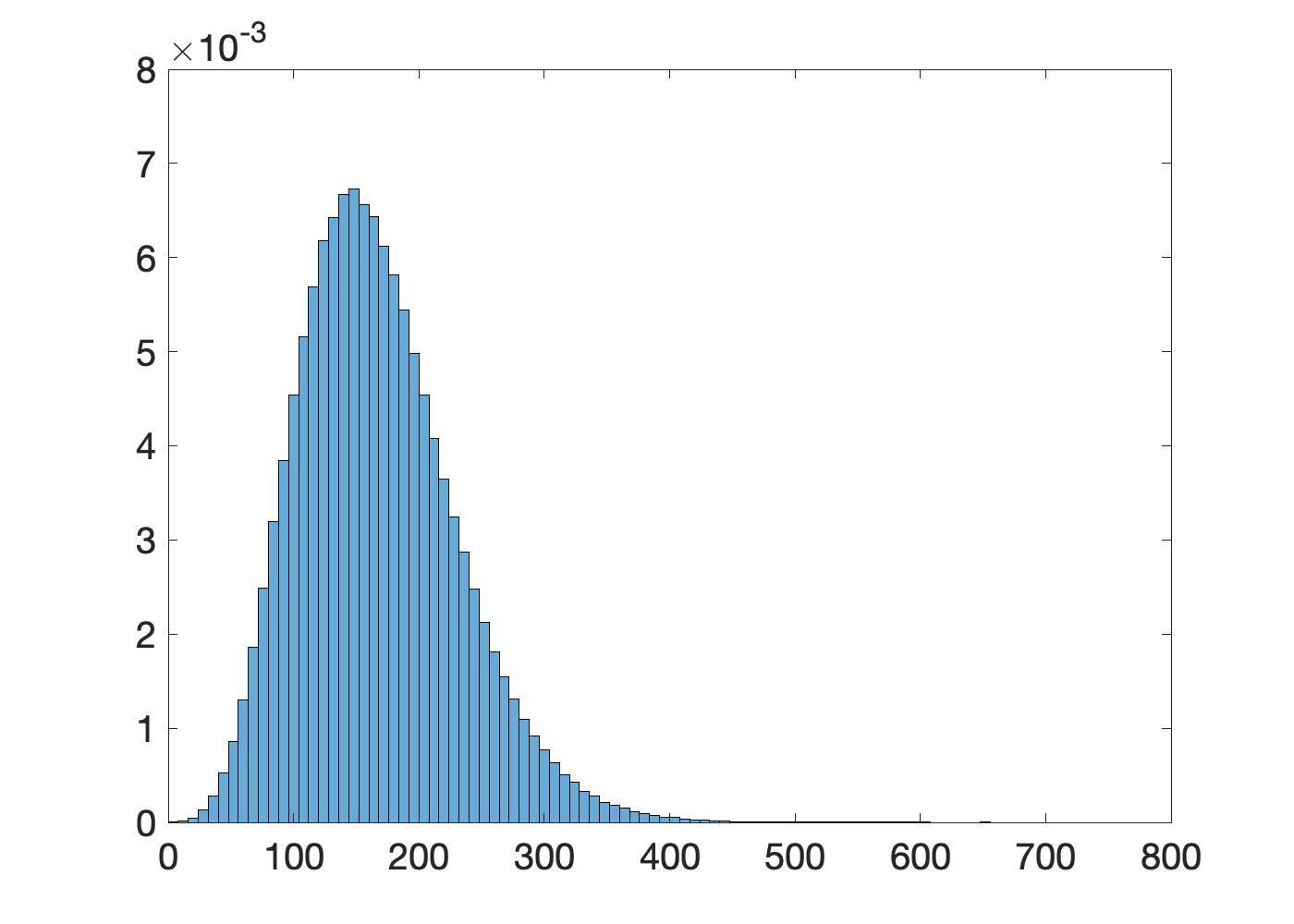}
         \caption{Tail Exponent: $-2$}
         \label{fig:1a}
     \end{subfigure}
     \begin{subfigure}[b]{0.4\textwidth}
         \centering
         \includegraphics[width=\textwidth]{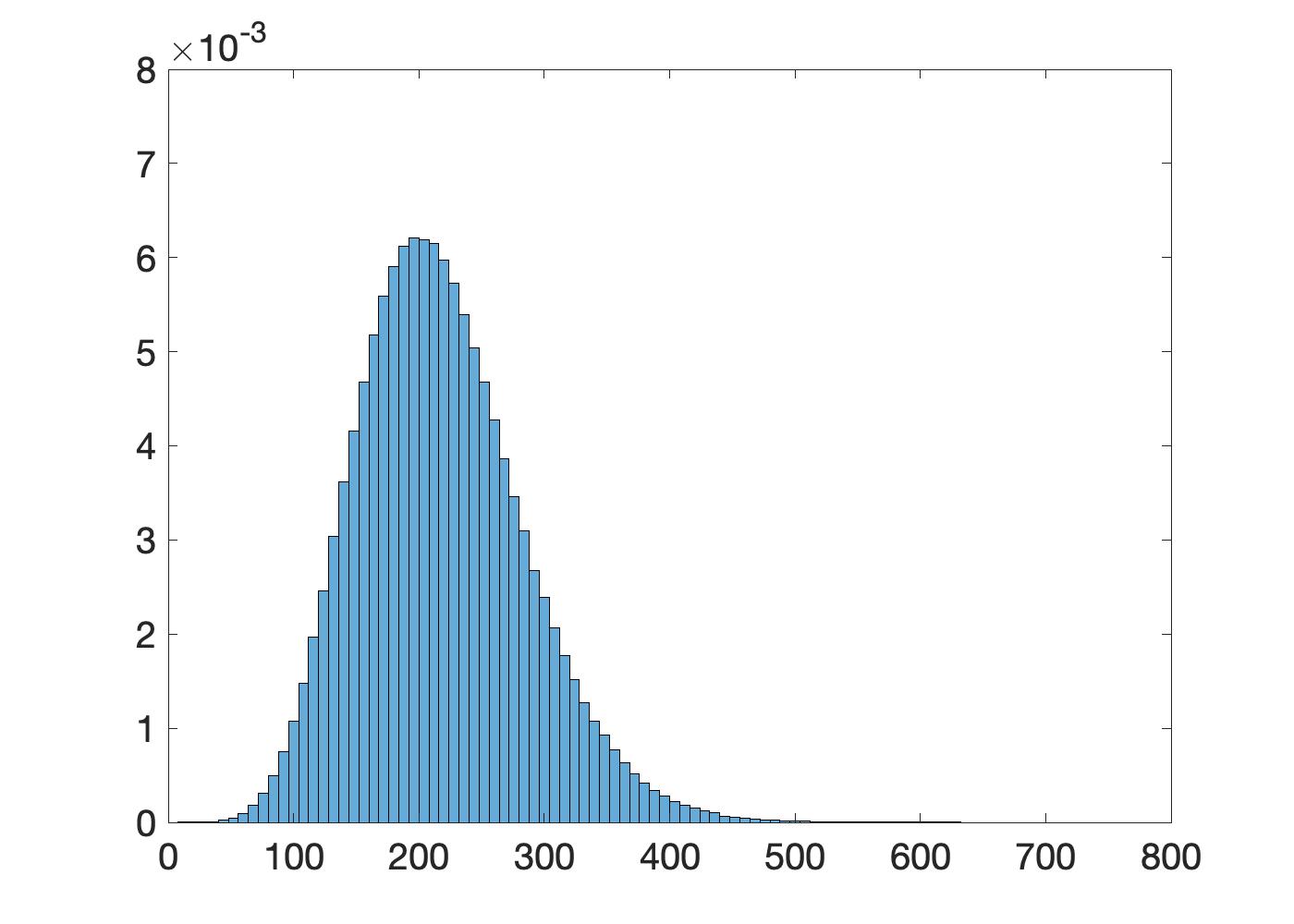}
         \caption{Tail Exponent: $-3$}
         \label{fig:1b}
     \end{subfigure}
     \vskip\baselineskip
    \begin{subfigure}[b]{0.4\textwidth}
         \centering
         \includegraphics[width=\textwidth]{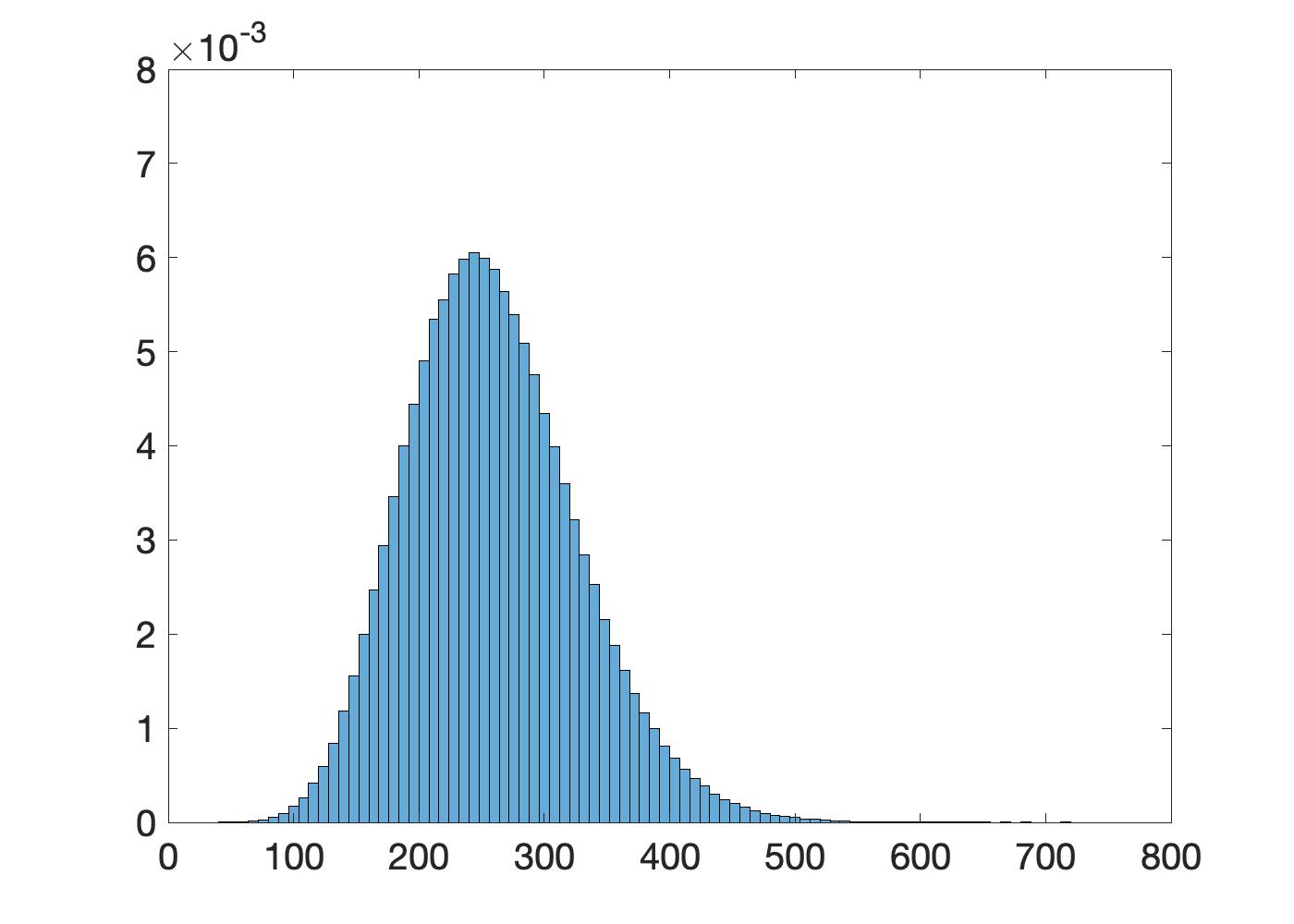}
         \caption{Tail Exponent: $-4$}
         \label{fig:1c}
     \end{subfigure}
     \begin{subfigure}[b]{0.4\textwidth}
         \centering
         \includegraphics[width=\textwidth]{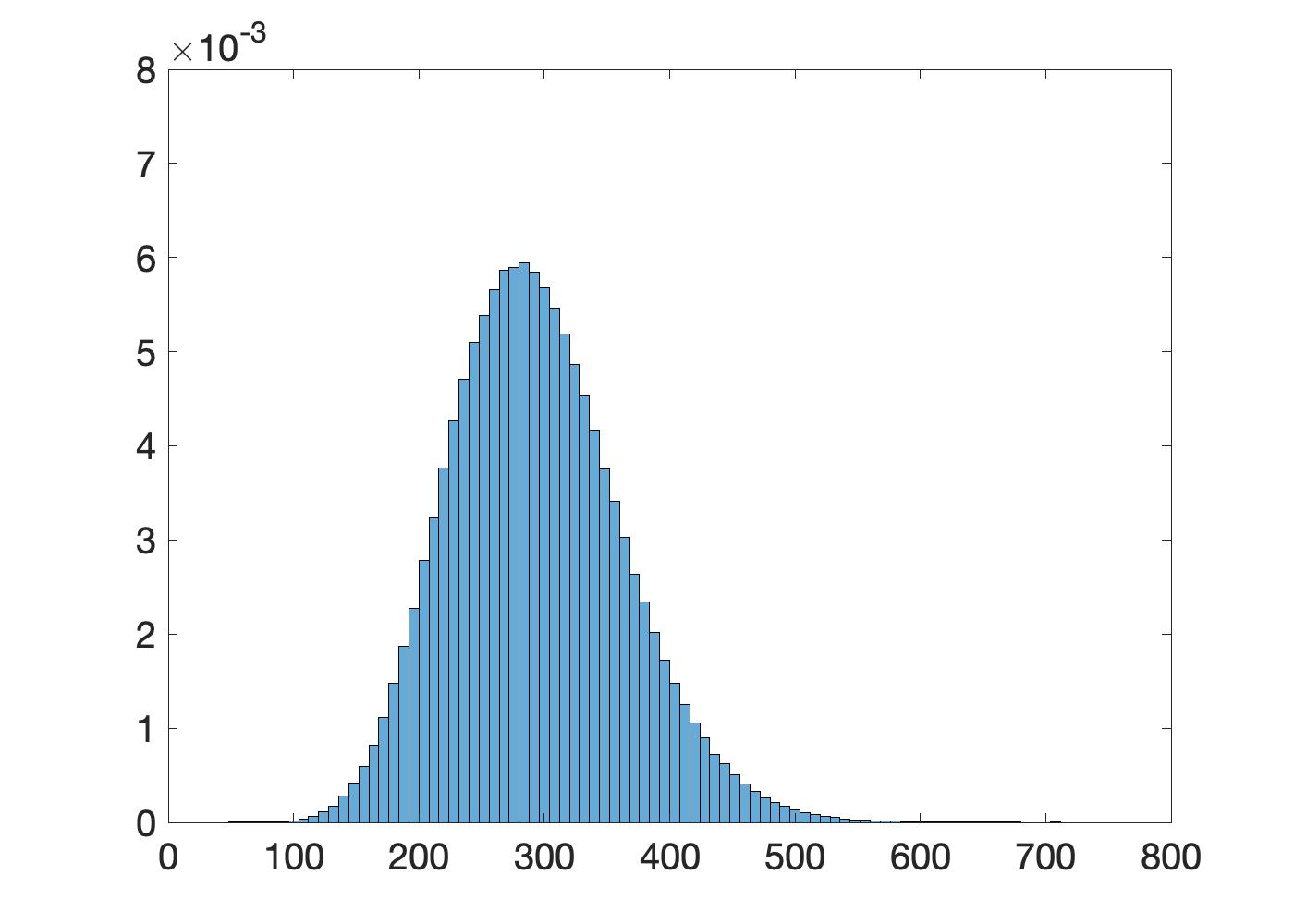}
         \caption{Tail Exponent: $-5$}
         \label{fig:1d}
     \end{subfigure}
        \caption{Distribution of the number of sub-optimal arm plays by UCB at time $T=2000$ for a two-armed Gaussian bandit with unit variances and arm mean gap $\Delta = 0.3$. In sub-figures (a)-(d), UCB is tuned to yield different tail exponents of the regret distribution. Each histogram consists of $10^6$ replications.}
        \label{fig:1}
\end{figure}

\begin{figure}
     \centering
     \begin{subfigure}[b]{0.4\textwidth}
         \centering
         \includegraphics[width=\textwidth]{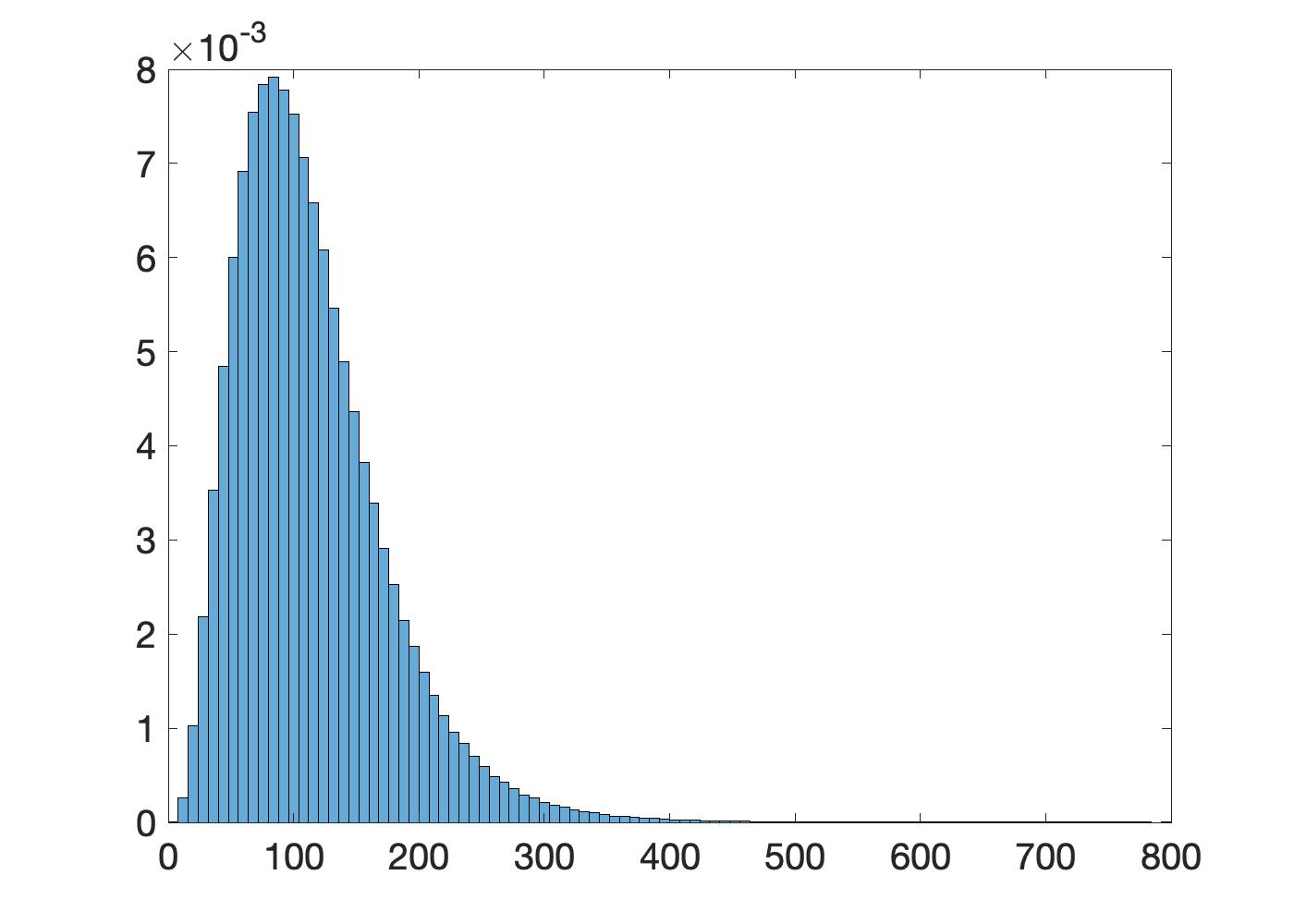}
         \caption{Tail Exponent: $-2$}
         \label{fig:2a}
     \end{subfigure}
     \begin{subfigure}[b]{0.4\textwidth}
         \centering
         \includegraphics[width=\textwidth]{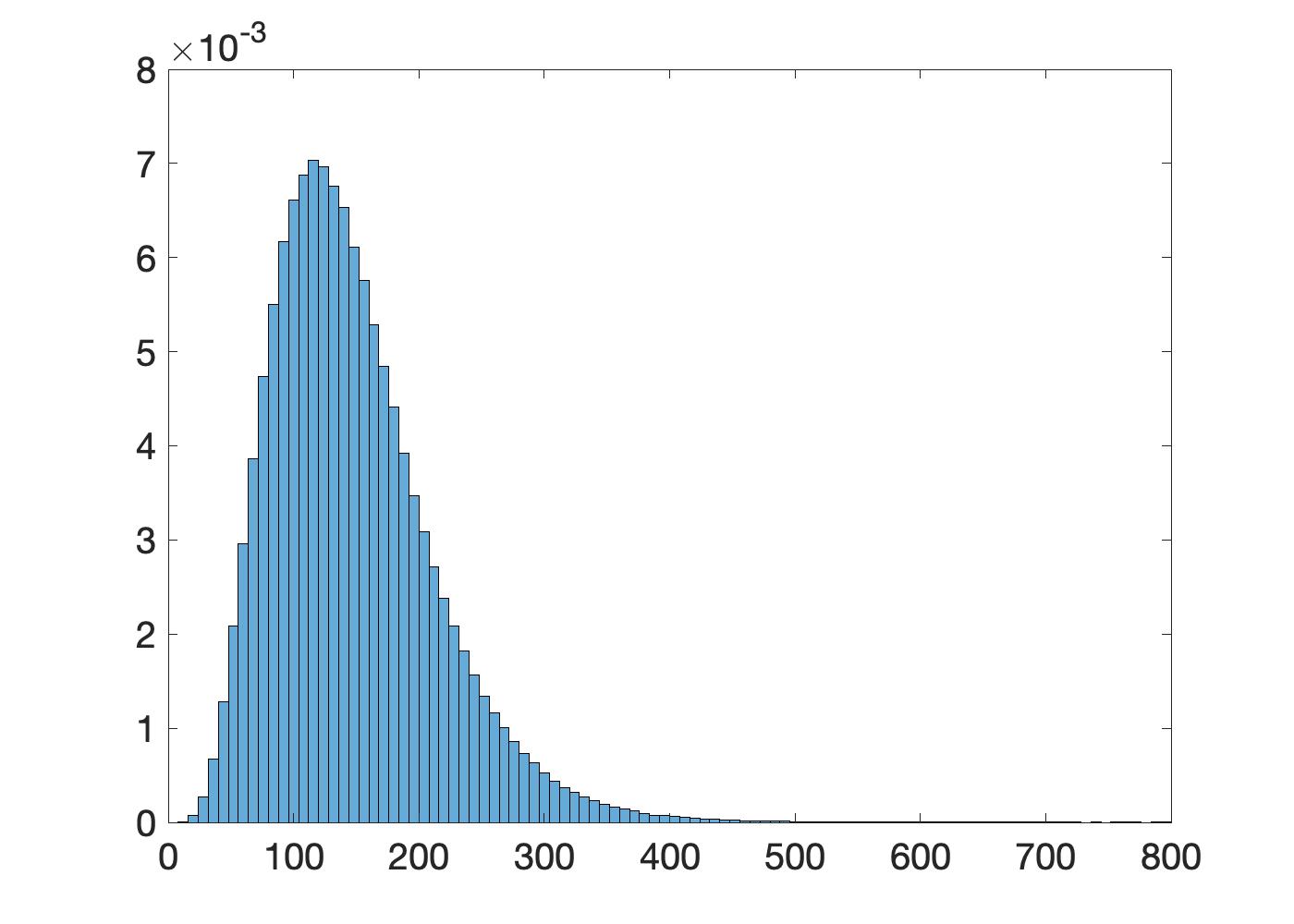}
         \caption{Tail Exponent: $-3$}
         \label{fig:2b}
     \end{subfigure}
     \vskip\baselineskip
    \begin{subfigure}[b]{0.4\textwidth}
         \centering
         \includegraphics[width=\textwidth]{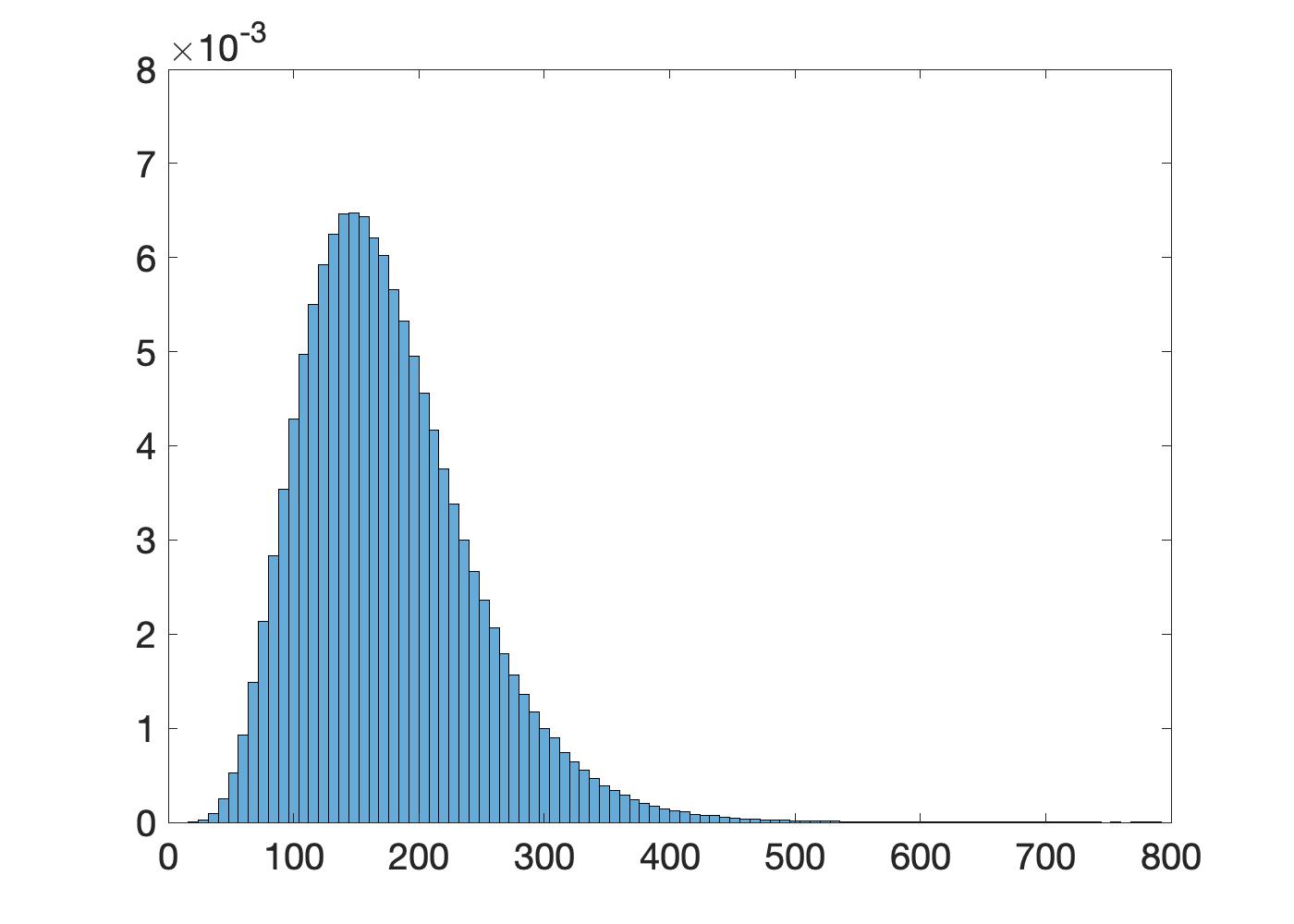}
         \caption{Tail Exponent: $-4$}
         \label{fig:2c}
     \end{subfigure}
     \begin{subfigure}[b]{0.4\textwidth}
         \centering
         \includegraphics[width=\textwidth]{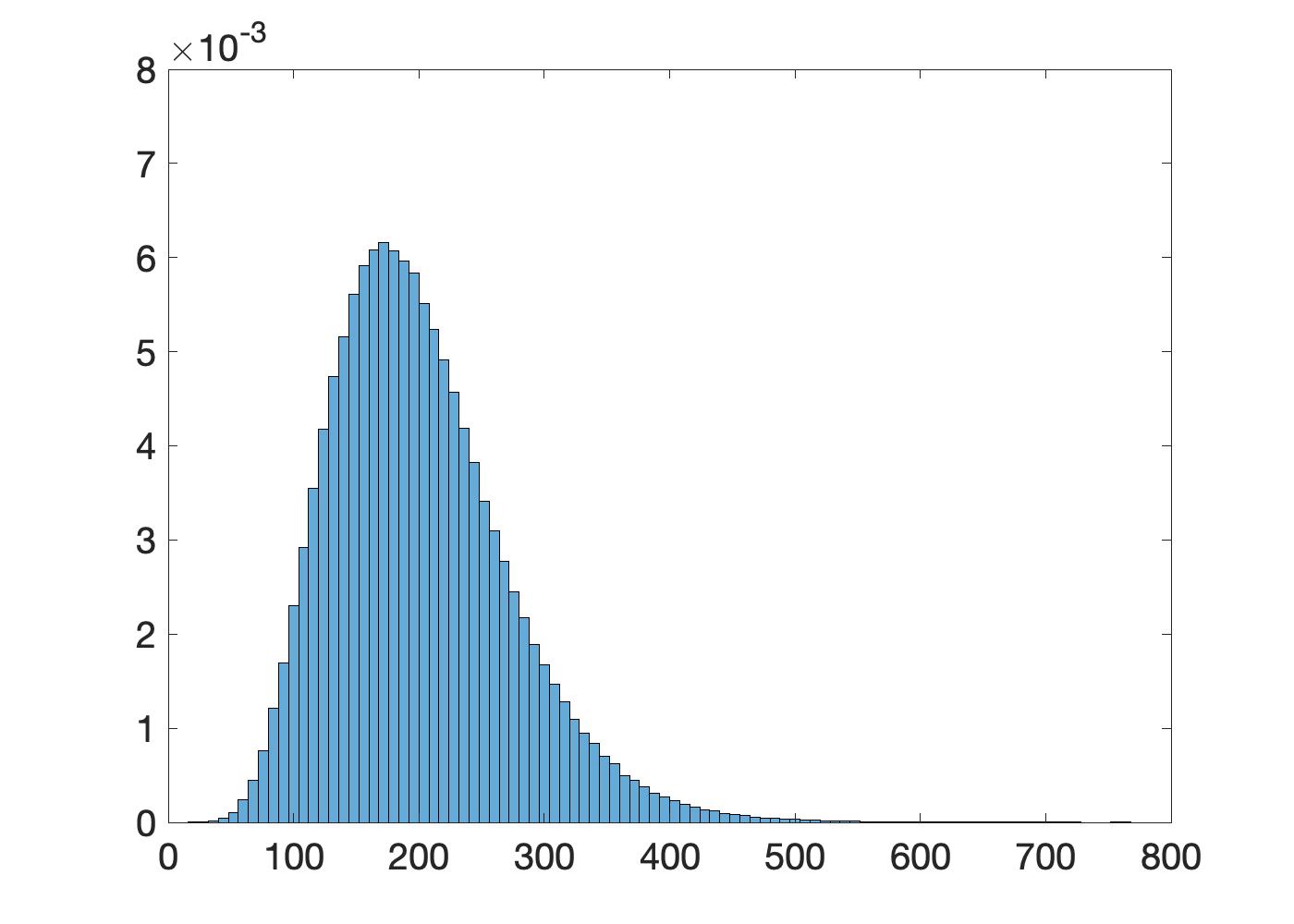}
         \caption{Tail Exponent: $-5$}
         \label{fig:2d}
     \end{subfigure}
        \caption{Distribution of the number of sub-optimal arm plays by TS at time $T=2000$ for a two-armed Gaussian bandit with unit variances and arm mean gap $\Delta = 0.3$. In sub-figures (a)-(d), TS is tuned to yield different tail exponents of the regret distribution. Each histogram consists of $10^6$ replications.}
        \label{fig:2}
\end{figure}

In this section, we numerically examine the CLT approximations of the regret of UCB and TS provided by Proposition \ref{prop2} (specifically (\ref{prop2_1})).
See Figures \ref{fig:1} and \ref{fig:2} for the UCB and TS (respectively) simulation results.
For both UCB and TS, we see that as the algorithms are modified so that the regret distribution tail is made lighter (i.e., by designing for rewards with larger variances, as discussed in Section \ref{mis}), the shape of the distribution becomes more like that of a Gaussian.
However, even when the regret tail is made lighter, the distributions in Figures \ref{fig:1} and \ref{fig:2} still exhibit some skewness (with a right tail).
This is more noticeable for TS, which has been empirically noted to exhibit more volatile regret behavior than UCB.
The regret of TS has more tendency to be at the extremes: either very low or quite high, thereby resulting in a more skewed regret distribution.

\begin{figure}
     \centering
     \begin{subfigure}[b]{0.49\textwidth}
         \centering
         \includegraphics[width=1.1\textwidth]{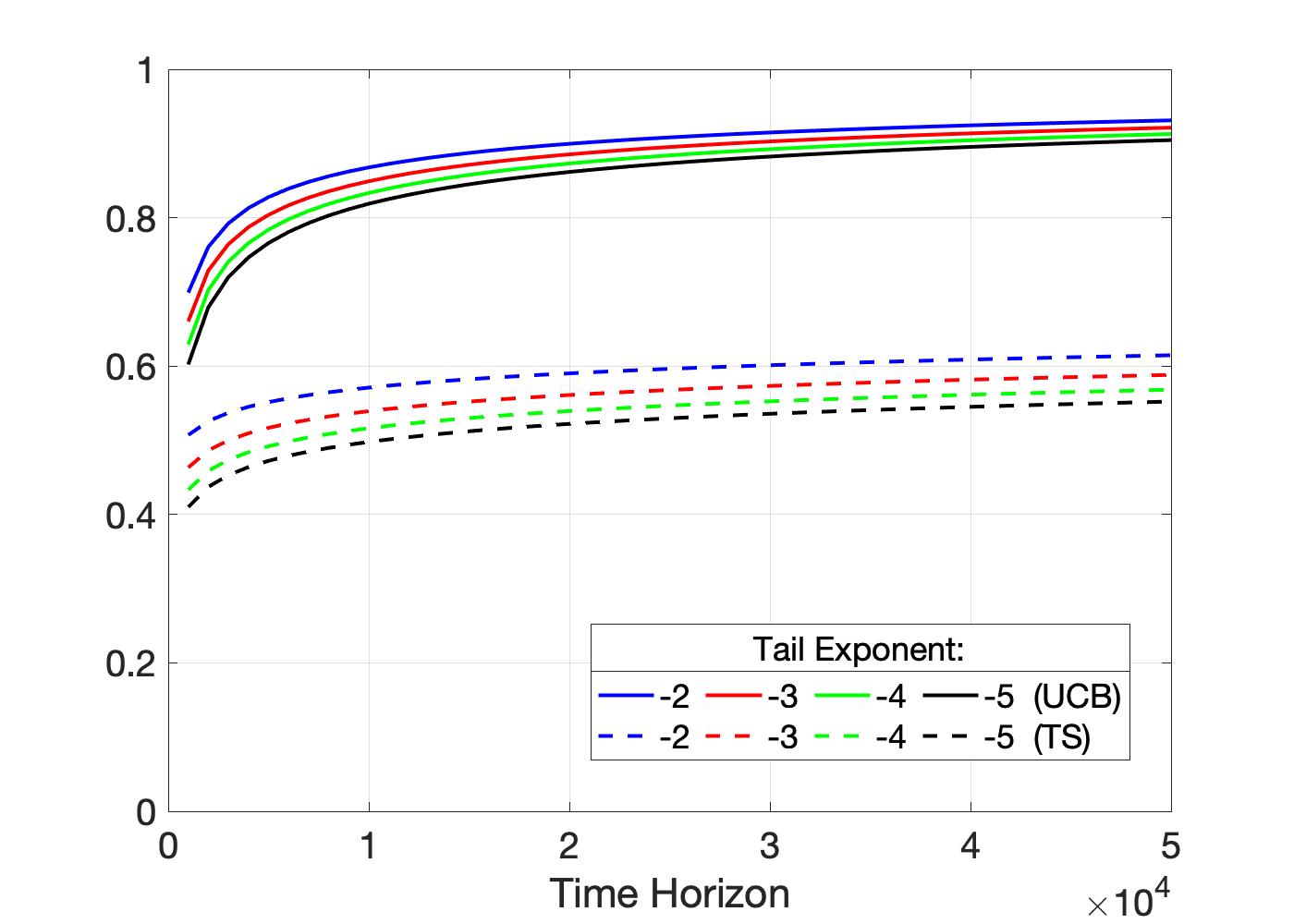}
         \caption{Observed Mean / CLT-predicted Mean}
         \label{fig:3a}
     \end{subfigure}
     \begin{subfigure}[b]{0.49\textwidth}
         \centering
         \includegraphics[width=1.1\textwidth]{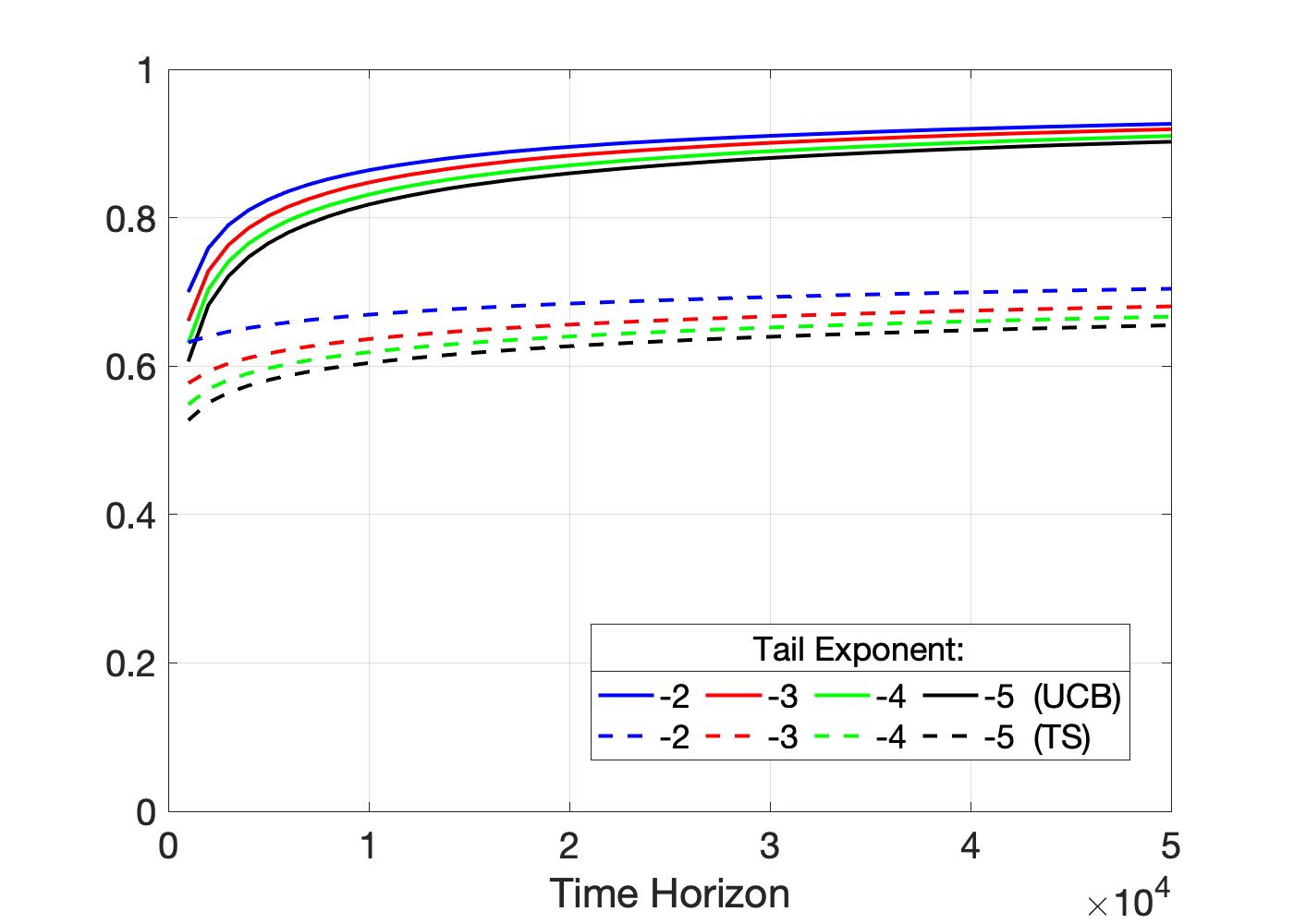}
         \caption{Observed Std Dev / CLT-predicted Std Dev}
         \label{fig:3b}
     \end{subfigure}
        \caption{Each curve tracks the ratio of the empirically-observed regret mean (standard deviation) to the CLT-predicted regret mean (standard deviation) over time. For different tail exponents (as indicated by the legend color), the curves for UCB are plotted using solid lines, while the curves for TS are plotted using dashed lines. In all cases, the environment is a two-armed Gaussian bandit with unit variances and arm mean gap $\Delta = 0.7$. Each curve is an average over $10^6$ replications.}
        \label{fig:3}
\end{figure}

In Figure \ref{fig:3}, we quantitatively examine the quality of the CLT approximation for the regret of UCB and TS.
In \ref{fig:3a}, we plot the ratio of the empirically-observed regret mean to the CLT-predicted regret mean.
In \ref{fig:3b} we plot the ratio of the empirically-observed regret standard deviation to the CLT-predicted regret standard deviation.
We see that the mean and standard deviation of regret predicted by the CLT are very good approximations for those of UCB.
However, the approximations for TS are poorer for the time horizons (up to $50,000$) included in the plots.
Nevertheless, the curves for TS are all monotone increasing, which suggests that the CLT approximation for the regret of TS keeps improving as the time horizon gets longer.

Interestingly, when we use versions of TS and UCB tuned to yield lighter regret tails (more negative tail exponents), it appears that longer time horizons are required for the ratios in Figures \ref{fig:3a} and \ref{fig:3b} to converge to $1$.
Nevertheless, we do find through simulations that for any fixed time horizon, the empirically-observed mean and standard deviation of regret corresponding to more negative tail exponents are strictly greater than those corresponding to less negative tail exponents.
This (perhaps somewhat obvious) qualitative finding agrees with the theory predictions in Propositions \ref{prop1} and \ref{prop2}.

\begin{APPENDICES}

\section{Technical Lemmas} \label{technical}

\begin{lemma} \label{lem000}
For any $z > 0$,
\begin{align*}
\frac{1}{4\sqrt{\pi z^2}} e^{-z^2/2} \le 1 - \Phi(z) \le \frac{1}{2\sqrt{\pi z^2}} e^{-z^2/2}.
\end{align*}
\end{lemma}

\proof{Proof of Lemma \ref{lem000}.}
See Formula 7.1.13 of \cite{abramowitz_etal1964}.
\halmos
\endproof

\begin{lemma} \label{lem00}
Let $a_1,a_2,\dots$ be a real-valued sequence. Then for any $n$,
\begin{align*}
    \max_{1 \le i \le n} a_i \le \log\left( \sum_{i=1}^n e^{a_i} \right) \le \max_{1 \le i \le n} a_i + \log(n).
\end{align*}
\end{lemma}

\proof{Proof of Lemma \ref{lem00}.}
The lower bound follows from:
\begin{align*}
    \max_{1 \le i \le n} a_i = \log\left( \max_{1 \le i \le n} e^{a_i} \right) \le \log\left( \sum_{i=1}^n e^{a_i} \right).
\end{align*}
The upper bound follows from:
\begin{align*}
    \log\left( \sum_{i=1}^n e^{a_i} \right) \le \log\left( n \cdot \max_{1 \le i \le n} e^{a_i} \right) = \max_{1 \le i \le n} a_i + \log(n).
\end{align*}
\halmos
\endproof

\begin{lemma} \label{lem0}
Let $p_j > 0$ be a sequence of probabilities such that $p_j \to 0$.
Let $G_j$ be a sequence of independent geometric random variables such that $G_j$ has corresponding success probability $p_j$.
Then for any $a > 0$,
\begin{align*}
    \frac{\max_{1 \le j \le n} \log(G_j p_j)}{n^a} \overset{\text{a.s.}}{\to} 0.
\end{align*}
\end{lemma}

\proof{Proof of Lemma \ref{lem0}.}
First, almost surely,
\begin{align}
    \liminf_{n \to \infty} \max_{1 \le j \le n} \log(G_j p_j) \ge 0. \label{lowerbound}
\end{align}
This follows from the fact that for $j$ sufficiently large,
\begin{align*}
    \P(G_j p_j < 1) \le 1 - (1-p_j)^{\lfloor p_j^{-1} \rfloor} \le 1 - 1/(2e).
\end{align*}
Next, by a straightforward argument,
\begin{align}
    \P\left(\max_{1 \le j \le n} \log(G_j p_j) > n^{a/2} \text{ i.o.}\right) = \P\left(\log(G_n p_n) > n^{a/2} \text{ i.o.}\right). \label{equivalence}
\end{align}
The right side of (\ref{equivalence}) is equal to zero by the Borel-Cantelli Lemma since
\begin{align*}
    \P\left(\log(G_n p_n) > n^{a/2}\right) = (1-p_n)^{\lfloor p_n^{-1} \exp(n^{a/2}) \rfloor} \le 2\exp(-\exp(n^{a/4}))
\end{align*}
for sufficiently large $n$.
Therefore, almost surely,
\begin{align}
    \limsup_{n \to \infty} \frac{\max_{1 \le j \le n} \log(G_j p_j)}{n^{a/2}} \le 1. \label{upperbound}
\end{align}
Together, (\ref{lowerbound}) and (\ref{upperbound}) give the desired result.
\halmos
\endproof

\end{APPENDICES}





\bibliographystyle{informs2014} 
\bibliography{references} 

\begin{thebibliography}{14}
\providecommand{\natexlab}[1]{#1}
\providecommand{\url}[1]{\texttt{#1}}
\providecommand{\urlprefix}{URL }

\bibitem[{Abramowitz \protect\BIBand{} Stegun(1964)}]{abramowitz_etal1964}
Abramowitz M, Stegun I (1964) \emph{{Handbook of Mathematical Functions with
  Formulas, Graphs, and Mathematical Tables}} (Dover).

\bibitem[{Auer et~al.(2002)Auer, Cesa-Bianchi, \protect\BIBand{}
  Fischer}]{auer_etal2002}
Auer P, Cesa-Bianchi N, Fischer P (2002) {Finite-time analysis of the
  multiarmed bandit problem}. \emph{Machine Learning} 47:235--256.

\bibitem[{Cowan \protect\BIBand{} Katehakis(2019)}]{cowan_etal2019}
Cowan W, Katehakis M (2019) {Exploration–exploitation policies with almost
  sure, arbitrarily slow growing asymptotic regret}. \emph{Probability in the
  Engineering and Informational Sciences} 1--23.

\bibitem[{Fan \protect\BIBand{} Glynn(2021{\natexlab{a}})}]{fan_etal2021a}
Fan L, Glynn P (2021{\natexlab{a}}) {Diffusion approximations for Thompson
  sampling}. \emph{arXiv:2105.09232} .

\bibitem[{Fan \protect\BIBand{} Glynn(2021{\natexlab{b}})}]{fan_etal2021b}
Fan L, Glynn P (2021{\natexlab{b}}) {The fragility of optimized bandit
  algorithms}. \emph{arXiv:2109.13595} .

\bibitem[{Gut(2009)}]{gut_2009}
Gut A (2009) \emph{{Stopped Random Walks: Limit Theorems and Applications}}
  (Springer).

\bibitem[{Kalvit \protect\BIBand{} Zeevi(2021)}]{kalvit_etal2021}
Kalvit A, Zeevi A (2021) {A closer look at the worst-case behavior of
  multi-armed bandit algorithms}. \emph{Advances in Neural Information
  Processing Systems} .

\bibitem[{Korda et~al.(2013)Korda, Kaufmann, \protect\BIBand{}
  Munos}]{korda_etal2013}
Korda N, Kaufmann E, Munos R (2013) {Thompson sampling for 1-dimensional
  exponential family bandits}. \emph{Conference on Neural Information
  Processing Systems} .

\bibitem[{Lai \protect\BIBand{} Robbins(1985)}]{lai_etal1985}
Lai T, Robbins H (1985) {Asymptotically efficient adaptive allocation rules}.
  \emph{Advances in Applied Mathematics} 6(1):4--22.

\bibitem[{Lattimore \protect\BIBand{} Szepesv\'ari(2020)}]{lattimore_etal2020}
Lattimore T, Szepesv\'ari C (2020) \emph{{Bandit Algorithms}} (Cambridge
  University Press).

\bibitem[{May et~al.(2012)May, Korda, Lee, \protect\BIBand{}
  Leslie}]{may_etal2012}
May B, Korda N, Lee A, Leslie D (2012) {Optimistic Bayesian sampling in
  contextual-bandit problems}. \emph{Journal of Machine Learning Research}
  13(1):2069--2106.

\bibitem[{Prabhu(1998)}]{prabhu_1998}
Prabhu N (1998) \emph{{Stochastic Storage Processes: Queues, Insurance Risk,
  Dams, and Data Communication}} (Springer).

\bibitem[{Thompson(1933)}]{thompson_1933}
Thompson W (1933) {On the likelihood that one unknown probability exceeds
  another in view of the evidence of two samples}. \emph{Biometrika}
  25(3):285--294.

\bibitem[{Wager \protect\BIBand{} Xu(2021)}]{wager_etal2021}
Wager S, Xu K (2021) {Diffusion asymptotics for sequential experiments}.
  \emph{arXiv:2101.09855v2} .

\end{thebibliography}

\end{document}